\documentclass[twoside,11pt]{article}

\usepackage[preprint]{jmlr2e}

\usepackage{amsmath}
\usepackage{amssymb}
\usepackage{bm}
\usepackage[capitalise]{cleveref}
\usepackage{wrapfig}
\usepackage{graphicx}
\usepackage{subfigure}
\usepackage[shortlabels]{enumitem}
\usepackage{xcolor}
\usepackage{natbib}

\usepackage{lastpage}
\jmlrheading{25}{2024}{1-\pageref{LastPage}}{1/24}{}{21-0000}{Alicia Curth, Alan Jeffares and Mihaela van der Schaar}

\ShortHeadings{Why do Random Forests Work?}{Curth, Jeffares and van der Schaar}
\firstpageno{1}

\begin{document}

\title{Why do Random Forests Work?\\ Understanding Tree Ensembles as Self-Regularizing Adaptive Smoothers}

\author{\name Alicia Curth \email amc253@cam.ac.uk \\
       \addr University of Cambridge
       \AND
       \name Alan Jeffares \email aj659@cam.ac.uk \\
       \addr University of Cambridge
       \AND
       \name Mihaela van der Schaar \email mv472@cam.ac.uk \\
       \addr University of Cambridge}

\editor{My editor}

\maketitle

\begin{abstract}%   <- trailing '%' for backward compatibility of .sty file
Despite their remarkable effectiveness and broad application, the drivers of success underlying ensembles of trees (especially random forests and gradient boosting) are still not fully understood. In this paper, we highlight how interpreting tree ensembles as adaptive and self-regularizing smoothers can provide new intuition and deeper insight to this topic.  We use this perspective to show that, when studied as smoothers -- whose predictions can be understood as simple weighted averages of the training labels -- randomized tree ensembles not only make predictions that are quantifiably more smooth than the predictions of the individual trees they consist of, but also further regulate their smoothness at test-time based on the dissimilarity between testing and training inputs. First, we use this insight to revisit, refine and reconcile two recent explanations of forest success by providing a new way of quantifying the conjectured behaviors of tree ensembles objectively by measuring the effective degree of smoothing they imply. Then, we move beyond existing explanations for the mechanisms by which tree ensembles improve upon individual trees and challenge the popular wisdom that the superior performance of forests should be understood as a consequence of variance reduction alone. We argue that the current high-level dichotomy into bias- and variance-reduction prevalent in statistics is insufficient to understand tree ensembles  -- because the prevailing definition of bias does not capture differences in the expressivity of the hypothesis classes formed by trees and forests. Instead, we show that forests can improve upon trees by three distinct mechanisms that are usually implicitly entangled. In particular, we demonstrate that the smoothing effect of ensembling can reduce variance in predictions due to noise in outcome generation, reduce variability in the quality of the learned function given fixed input data \textit{and} reduce potential bias in learnable functions by enriching the available hypothesis space.

\end{abstract}


\section{Introduction}
Random forests \citep{breiman2001random} have emerged as one of the most reliable off-the-shelf supervised learning algorithms \citep{fernandez2014we} and remain among the most popular and successful methods in data science applications \citep{mentch2020randomization}, succeeding on tabular data in particular \citep{grinsztajn2022tree}. Beyond their original use in standard prediction settings, they have since also been successfully extended to other more involved problems, such as quantile regression \citep{meinshausen2006quantile}, survival analyses \citep{hothorn2006survival, ishwaran2008random} and causal inference \citep{wager2018estimation, athey2019generalized}. Despite their robust out-of-the-box performance across all such applications, only a few recent attempts exist at gaining a deeper understanding of the drivers of their success \citep{wyner2017explaining, mentch2020randomization, zhou2023trees}. In this work, we aim to build upon these recent efforts to provide intuitive explanations for forest performance by exploring connections of tree ensembles with the classical statistics literature on \textit{regression smoothers} \citep{hastie1990GAM}.

\paragraph{Related work.} The term Random Forest as used today usually refers to \cite{breiman2001random}'s implementation of a randomized tree ensemble; however, a much broader methodological literature on tree ensembles (or forests) led to the development and popularization of such algorithms (e.g. \citet{ho1995random,breiman1996bagging, amit1997shape, ho1998random, dietterich2000experimental, breiman2000randomizing, cutler2001pert}). Since then, there have been many efforts to better understand the statistical properties of tree ensembles, e.g. their consistency \citep{biau2008consistency, arnould2023interpolation}, also by studying them as kernel methods \citep{biau2010layered, davies2014random, scornet2016random, olson2018making}, confidence intervals \citep{wager2014confidence} and algorithmic convergence \citep{lopes2020measuring}. For an in-depth overview of research on tree ensembles, refer to \cite{biau2016random}.

Surprisingly, however, relatively little work has been dedicated to providing intuitive explanations for the success of tree-based ensembles in practical applications. Instead, the success of random forests relative to individual trees is usually discussed only at a high-level and attributed to variance reduction effects \textit{alone} \cite[Ch. 15]{hastie2009elements}. In this paper, we focus on two recent and much more specific explanations for forest success. A first novel perspective was presented in \cite{wyner2017explaining} who conjecture that \textit{interpolating} tree ensembles (i.e. ensembles achieving zero training error, including both random forests and boosted ensembles) succeed due to having a ``spiked-smooth'' property, where model averaging ensures that overfitting to training points only happens locally. A second explanation was then put forward by \cite{mentch2020randomization} (and expanded on in \cite{zhou2023trees}), who conjecture that the randomness in the forest-construction process has regularizing properties, and show that randomness in split-feature selection reduces a measure of \textit{the degrees of freedom} used by a model \citep{efron1986biased, tibshirani2015degrees} -- which is why \cite{mentch2020randomization} argue that random forests work particularly well in settings with low signal-to-noise ratio. 

\begin{figure}[!t]
    \centering
    \vspace{-0.25cm}
\includegraphics[width=.99\textwidth]{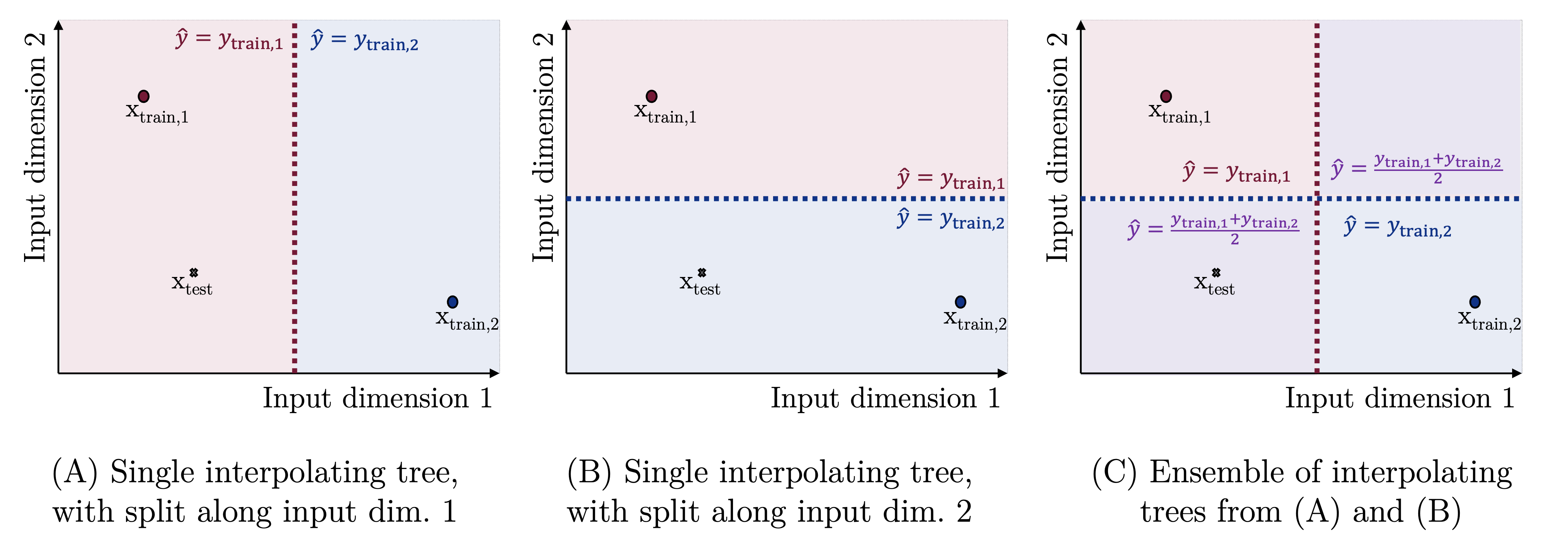}\vspace{-0.25cm}
    \caption{\textbf{Illustrating when, where and why averaging ensembles can make smoother predictions than individual regression trees.} \small Consider a stylized example with two input-output pairs $(x_{train, i}, y_{train,i})$ to which we fit decision trees of full depth to learn to make predictions $\hat{y}$. This problem is \textit{underspecified} -- the splits with decision boundaries displayed in panels (A) and (B) lead to \textit{the same} impurity decrease. Which split will be realized is thus a random choice. Both trees make \textit{the same prediction} for training inputs $x_{train, i}$. For the previously unseen test-input $x_{test}$, however, the two trees will \textit{issue different predictions}. An ensemble of these trees, as displayed in Panel (C), will therefore (i) make the same prediction as the individual trees close to training examples but (ii) will make a \textit{smoother} prediction than each individual tree in regions where the decision boundary is underdetermined. That is, while individual full-depth trees always act as 1-Nearest Neighbor estimators (with learned distance metric), an ensemble of 2 full-depth trees will act as a 1-Nearest Neighbor estimators around training examples, but can self-regularize to act as a 2-Nearest Neighbor estimator in underdetermined regions at test-time.}
    \label{fig:page1fig}
    \vspace{-.3cm}
\end{figure}

\paragraph{Outlook.} In this work, we contribute new insight to this discussion by relying on the \textit{adaptive nearest neighbor} interpretation of tree ensembles: Trees can also be interpreted as adaptive nearest neighbor estimators with a learned distance metric, and ensembles of such trees are then adaptive weighted nearest neighbor estimators \citep{lin2006random}.  Similar to \cite{curth2023u}, who studied the appearance of double descent in tree ensembles in \cite{belkin2019reconciling} through this lens, we use this perspective to link the literature on understanding the success of tree ensembles to the literature on \textit{smoothers} \citep{hastie1990GAM} in \cref{sec:smooth-back}. This allows us to build new intuition as we can now study the predictions of trees and ensembles thereof -- which are highly complex objects to describe otherwise -- as simple weighted averages of the training labels, enabling us to investigate and quantify the \textit{effective amount of smoothing} performed by such models. In particular, as illustrated in \cref{fig:page1fig}, we use this perspective to demonstrate that (and why) randomized ensembles often make predictions that are not only quantifiably more smooth than the predictions of the individual trees they consist of, but also more smooth at test-time when given previously unseen inputs than when given training inputs. 

In the first part (\cref{sec:reconcile}), this insight allows us to revisit, refine and reconcile the two recent explanations of forest success discussed above by \textit{quantifying} the conjectured behaviors of tree-ensembles and the effect of hyperparameters objectively. In \cref{sec:wynersec}, we first show that the \textit{``spiked-smooth''} behavior of interpolating tree ensembles conjectured by \cite{wyner2017explaining} can be \textit{quantified} through the gap in effective parameters used by forests when issuing predictions at training inputs versus at previously unseen testing inputs. Going beyond \cite{wyner2017explaining}'s original interpolation argument, we also demonstrate that this input-dependent difference in the amount of smoothing applied is \textit{not} actually unique to forests that can interpolate the training data perfectly; yet, it appears to be more pronounced the more overfitted individual models are to the training data. In \cref{sec:mentchsec}, we then show that the degrees of freedom measure studied by \cite{mentch2020randomization} falls short in explaining some important performance differences of forests relative to trees -- essentially because it measures only the dependence of training predictions on their own training label, but not how much smoothing is applied across \textit{other labels} -- and that this can be remedied by using the effective parameter measure studied in \cite{curth2023u} instead. Finally, in \cref{sec:recsec}, we argue that while \cite{mentch2020randomization} imply that the two explanations of forest success are competing, \cite{wyner2017explaining}'s ``spiked-smooth interpolation'' explanation is actually better understood as a special case of their ``randomization as regularization'' explanation: we show that ``spiked-smooth'' behavior is in fact \textit{a result} of the smoothing regularization induced by randomization in forests. 

In the second part (\cref{sec:bias-var}), we then move beyond existing explanations and investigate the roles of bias and variance-reduction in determining the superior test-time performance of random forests relative to individual trees.  In particular, in \cref{sec:theorybiasvar} we challenge the prevailing wisdom that ensembles \textit{only} improve over individual trees due to variance-reduction effects by highlighting that this conclusion is largely based on a notion of bias that is insufficient to capture the driving forces behind forest success:  the definition for bias usually employed in the statistics literature cannot capture the difference in expressivity of the hypothesis classes formed by trees and forests. We then highlight that there are indeed differences in what functions can be \textit{represented} by trees and forests, which can lead to another form of bias differing between the two methods. Thus, we argue that forests improve upon trees by three separate mechanisms that are usually implicitly entangled: the smoothing effect achieved by ensembling can reduce variance in predictions due to noise in outcome generation, reduce variability in the quality of the learned function given fixed input data \textit{and} reduce potential bias in learnable functions by enriching the available hypothesis space. In \cref{sec:emp-bias-var}, we then investigate these mechanisms empirically. We first confirm in \cref{sec:snr} that when considering in-sample prediction error, i.e. test performance on inputs previously observed during training, ensembles improve upon individual trees mainly because they reduce the impact of noise in outcomes. However, we then show in \cref{sec:bias} that when generalization to \textit{unseen} inputs is of interest, forests are indeed able to \textit{realize} different predictions than the individual trees they consist of, which constitutes \textit{another important reason} they can perform better for testing points further from the support of the training data.

\section{Understanding Tree Ensembles as Adaptive Smoothers}\label{sec:smooth-back}

In this section, we focus on formalizing and understanding trees and forests as adaptive smoothers. We begin with a general introduction to regression smoothers (\cref{sec:background}), after which we formalize trees (\cref{sec:treesassmoother}) and tree ensembles (\cref{sec:forestassmoother}) in this framework. We then discuss how one can quantify the effective degree of smoothing performed by such smoothers (\cref{sec:eff-p-measure}) and close this section by discussing why train- and test-time behavior of trees and forests are likely to differ (\cref{sec:effp-diff}). 

\subsection{Background on smoothers}\label{sec:background}
Smoothers are a class of supervised learning methods that learn to predict outcomes $Y\!\in\! \mathcal{Y}\! \subset\! \mathbb{R}$ from  inputs $X\! \in\! \mathcal{X} \subset \mathbb{R}^d$ by averaging -- i.e. ``smoothing'' -- over values of $Y$ observed in the training data (see e.g. \citet[Ch. 2]{hastie1990GAM}). Let $\mathcal{D}^{\text{train}}\!=\! \{(x_i, y_i)\}^n_{i=1}$ denote the train-time realizations of $(X, Y) \!\in\! \mathcal{X} \!\times\! \mathcal{Y}$ and $\mathbf{y}_{\text{train}}\! =\!(y_1, \! \ldots\!, y_n)^T$ the $n\times 1$ vector of training outcomes with indices \smash{$\mathcal{I}_{\text{train}}\!=\!\{1, \ldots, n\}$}. Then, for some input \smash{$x_0 \!\in \! \mathcal{X}$}, smoothers issue predictions that are linear combinations of the training labels:
\begin{equation}\label{eq:smootherdef}
\textstyle \hat{f}(x_0)=\hat{\mathbf{s}}(x_0) \mathbf{y}_{\text{train}}  = \sum_{i \in \mathcal{I}_{\text{train}}} \hat{s}^i(x_0) y_i
\end{equation}
where the $1\! \times\! n$ vector $\hat{\mathbf{s}}(x_0)= (\hat{s}^1(x_0), \ldots, \hat{s}^n(x_0))$ contains smoother weights for input $x_0$. Prototypical examples of smoothers rely on weighted averages (most prominently used for scatterplot smoothing in the $d=1$ case \citep[Ch. 2.2]{hastie1990GAM}); well-known members of this class of prediction methods include k-nearest neighbor (k-NN) methods, kernel smoothers and even linear regression \cite[Ch. 2-3]{hastie1990GAM}. While smoothers are usually used in a regression context, \cref{eq:smootherdef} could also be used to predict class probabilities in a classification context instead.

The smoothers traditionally considered in the statistics literature are both \textit{linear} and \textit{fixed}. A smoother is \textit{linear}  if  $\hat{\mathbf{s}}(\cdot)$ does not depend on $\mathbf{y}_{\text{train}}$, i.e. if the smoother weights will be the same regardless of the observed training outcome values. We consider a smoother \textit{fixed} (or non-random) if $\hat{\mathbf{s}}(\cdot)$ is \textit{deterministic} given observed training data. Both are natural properties of prototypical smoothers: kernel weights, nearest neighbor distances and projection matrices -- which determine  $\hat{\mathbf{s}}(\cdot)$ for kernel smoothers, k-NN estimators and ordinary least squares regression, respectively -- are independent of training outcomes and fixed given the training input points.

\subsection{Trees as adaptive smoothers}\label{sec:treesassmoother}
In this paper, we exploit that trees are sometimes considered \textit{adaptive} nearest neighbor methods \citep{lin2006random}, i.e. nearest neighbor smoothers with an adaptively constructed (outcome-oriented) kernel (their implied kernels have also been the subject of theoretical study in e.g. \citet{davies2014random, scornet2016random, olson2018making}). This is because regression trees (as well as classification trees that make predictions through averaging, but not voting\footnote{This restriction to \textit{averaging} trees is similar to e.g. \cite{wager2018estimation}, and is supported by recent work suggesting using the squared loss for classification \citep{hui2020evaluation, muthukumar2021classification}.}), issue predictions that are simply an average across the outcomes of training instances falling within the same terminal leaf -- which is of the form of \cref{eq:smootherdef} where the weights $\hat{\mathbf{s}}(x_0)$ are \textit{learned} and hence depend on the training outcomes. 

To make this more formal, define trees as predictors $T_\Theta(\cdot) \equiv T(\cdot; \Theta): \mathcal{X} \rightarrow \mathbb{R}$. As in \citet[Ch. 15]{hastie2009elements} we use the random variable $\Theta$ to fully characterize the tree in terms of its split variables, cutpoints and terminal leaf values. The value of this random variable $\Theta$ usually depends on (i) the random training sample $Z=\{(X_i, Y_i)\}^n_{i=1}$ (we drop explicit dependence of $\Theta \equiv \Theta(Z)$ on $Z$ in the remainder for notational convenience) and (ii) additional randomness in the tree-building procedure which determines $\mathbb{P}(\Theta| Z)$, the distribution over possible trees given the training data\footnote{We remain agnostic to how trees are actually built, e.g. using the CART criterion \citep{breiman1984classification}.}. Then, note that the predictions of any regression tree with $n_{leaf}$ leaves can be represented by $n_{leaf}$ contiguous axis-aligned hypercubes $L_{\Theta}^p \in \mathcal{L}$, each associated with the mean outcome of all training examples that fall into this terminal leaf. Thus, if we let $l_{\Theta}(\cdot): \mathcal{X} \rightarrow \mathcal{L}$ map input $x_0$ to its corresponding leaf and  let $n_{l_{\Theta}(x_0)}=\sum^n_{i=1} \bm{1}\{l_{\Theta}(x_0)=l_{\Theta}(x_i)\}$ denote the number of training examples in this leaf, we can write the predictions of a tree as
\begin{equation}\label{eq:tree-pred}
 \textstyle T_\Theta(x_0)= \sum_{i \in \mathcal{I}_{\text{train}}} \frac{\bm{1}\{l_{\Theta}(x_0)=l_{\Theta}(x_i)\}}{n_{l_{\Theta}(x_0)}}y_i  
\end{equation}
By comparing \cref{eq:tree-pred} with \cref{eq:smootherdef}, it becomes clear that we can write a tree's prediction in smoother form with smoother weights
 \begin{equation}\label{eq:s_tree}
     \hat{\mathbf{s}}_{\Theta}(x_0) = \left(\frac{\bm{1}\{l_{\Theta}(x_0)=l_{\Theta}(x_1)\}}{n_{l_{\Theta}(x_0)}}, \ldots, \frac{\bm{1}\{l_{\Theta}(x_0)=l_{\Theta}(x_n)\}}{n_{l_{\Theta}(x_0)}}\right)
 \end{equation}

\paragraph{How do trees differ from classical smoothers?} Trees differ from classical smoothers because they are generally both \textit{non-linear} (or: adaptive) and \textit{random}. While trees can be \textit{expressed} in linear form, they are usually\footnote{This does not hold for tree building procedures that do not use $\mathbf{y}_{\text{train}}$, like totally randomized trees \citep{geurts2006extremely}; these are linear smoothers.} not actually linear smoothers because $\hat{\mathbf{s}}_{\Theta}(\cdot)$ depends on $\mathbf{y}_{\text{train}}$ through ${\Theta(Z)}$ -- they are thus \textit{outcome-adaptive}. Further, the smoother weights implied trees are also usually \textit{not fixed} but random -- given training data, $\Theta$ is usually a random variable over multiple possible tree structures as the regression problem is generally underspecified. For example, the choice between multiple split variables that all bring \textit{the same purity increase} is random in popular implementations (e.g. in \texttt{sklearn}'s \citep{scikit-learn} implementation of trees where ties are broken by choosing the variable that appears first in a random ordering). Some properties studied in the classical statistics literature on smoothers therefore do \textit{not} carry over to tree-based smoothers directly: e.g. the well-known formulas for predictive variance and influence of individual training observations (see e.g. \citet[Ch. 3]{hastie1990GAM})  no longer apply immediately. This is not problematic for the purposes of this paper, as we are interested in studying trees (and ensembles thereof) as smoothers due to other properties that do hold up in this context.  Essentially, the smoother perspective enables more intuitive insights because we can now study the predictions of trees and forests -- which are highly complex objects to describe otherwise -- as simple weighted averages of the training labels. In particular, this allows us to i) make use of tools for \textit{quantifying} the effective degree of smoothing performed by them, as we further discuss in \cref{sec:eff-p-measure} and ii) much more easily reason about their expected behavior as in e.g. \cref{sec:effp-diff}. 

\subsection{Tree ensembles as adaptive smoothers}\label{sec:forestassmoother}
 An \textit{ensemble} \citep[see e.g.][]{dietterich2000ensemble} of $B$ models $\hat{f}_b(\cdot)$ with ensemble-weights $w_b$ issues predictions that are averaged across its individual members as
\begin{equation}\label{eq:ens}
\textstyle    \hat{f}^{ens}(x_0) = \sum^B_{b=1}w_b  \hat{f}_b(x_0)
\end{equation}
Unless indicated otherwise, we will generally use the term \textit{ensemble} synonymously with the standard case where $w_b=\frac{1}{B}$. Further, note that it is considered desirable to have some \textit{diversity} between predictions of the different ensemble members \citep{krogh1994neural} as otherwise the prediction of the ensemble will be the same as the predictions of the individual (near-) identical ensemble members.

 It is easy to see that an ensemble of $B$ smoothers with smoother weights $\hat{\mathbf{s}}_{b}(x_0)$ will itself be a smoother with weights $\hat{\mathbf{s}}_{ens}(x_0)=\sum^B_{b=1}w_b \hat{\mathbf{s}}_{b}(x_0)$ as its predictions can be written as
\begin{equation}
     \hat{f}^{ens}(x_0) = \sum^B_{b=1}w_b  \hat{f}_b(x_0) = \sum^B_{b=1}w_b \hat{\mathbf{s}}_{b}(x_0)\mathbf{y}_{\text{train}}= \hat{\mathbf{s}}_{ens}(x_0)\mathbf{y}_{\text{train}}
\end{equation}

Ensembles of trees -- also known as forests -- then issue predictions \newline $    {F}_{B, \mathbf{\Theta}}(x_0) =  \sum^B_{b=1} w_b T(x_0; \Theta^b)$ by averaging across trees created through $B$ different random realizations of $\Theta^b$, collected in $\mathbf{\Theta}=(\Theta^1, \ldots, \Theta^B)$. Thus, a forest of size $B$ is a smoother with weights $\hat{\mathbf{s}}_{B, \mathbf{\Theta}}(x_0)= \sum^B_{b=1} w_b\hat{\mathbf{s}}_{\Theta^b}(x_0)$.

\paragraph{Inducing variation across trees in forests.} While \textit{some} randomness exists in tree-building processes naturally due to underdetermination (as discussed above), many different types of tree ensembles have been proposed in the literature to induce \textit{additional} variation across ensemble members. Most prominently, ensembling through \textit{bootstrap aggregation} \citep{breiman1996bagging} (a.k.a. \textit{bagging}) constructs individual ensemble members from different bootstrap samples of the original training data\footnote{In this case, the smoother weights in individual trees are not given by \cref{eq:s_tree} exactly but augmented with a counter for how often any individual observation appears in a bootstrap sample.}, and ensembling through \textit{feature subsampling} induces randomness by varying the subset of the $d$ input features used in building the individual ensemble members. In the context of trees, the latter can be achieved either by randomly sampling the features available for constructing the full trees (i.e. subspacing, as proposed in \citet{ho1998random}), or by randomly sampling the features available for constructing each split  (i.e. random feature selection, as proposed in \cite{amit1997shape}). Although a wide range of randomized tree ensembles have historically been referred to as random forests (see e.g. \cite{lin2006random}), the term today appears to be synonymous with \cite{breiman2001random}'s Random Forest -- which is a randomized tree ensemble that uses \textit{both} bootstrapping and random subsampling of $p_{try} \leq d$ input features at each split (see e.g. \cite[Ch. 15]{hastie2009elements}). Throughout this paper, we will consider both (i) $m=\frac{p_{try}}{d}$, the proportion of features considered at each split and (ii) the choice whether or not to use bootstrapping as \textit{hyperparameters} of random forest algorithms (in addition to (iii) the depth of the individual trees considered) and will investigate their effects individually.

\paragraph{Boosted ensembles.} In all tree ensembles discussed so far, individual ensemble members are trained independently of each other. In \textit{boosted} ensembles, however, members are trained \textit{sequentially} to iteratively improve upon the performance of the ensemble as a whole \citep{schapire2003boosting}. In \cref{app:boost}, we show that boosted tree ensembles created using gradient boosting \citep{friedman2001greedy} can also be expressed as smoothers, albeit with $w_b\neq \frac{1}{B}$. %Note that, because boosted ensembles do not generally use , their smoother weights will generally have $\sum^n_{i=1}\hat{\mathbf{s}}_{\mathbf{\Theta}^B(Z)}^i(x_0)\neq 1$.

\subsection{Quantifying the effective degree of smoothing in smoother predictions}\label{sec:eff-p-measure}
Our main motivation for studying trees (and ensembles thereof) as smoothers is that this makes it much easier to reason about their predictions as weighted averages of training labels \textit{and} allows us to make use of tools for \textit{quantifying} the effective degree of smoothing performed by predictors -- which we will show to be instrumental to understanding their performance. Measures for the \textit{effective number of parameters} (also known as \textit{degrees of freedom}) used by a smoother were introduced in the smoothing literature to make different smoothers quantitatively comparable with respect to the amount of smoothing they do \cite[Ch. 3.5]{hastie1990GAM}. Most such measures summarise properties of the smoother when issuing predictions on the training data, i.e. consider summaries of $\{\hat{\mathbf{s}}(x_i)\}_{i \in \mathcal{I}_{\text{train}}}$, because this allows to quantify dependence of predictions at train observation points $x_i$ on \textit{their own} training labels $y_i$. Intuitively, this captures the potential for overfitting on noise in training realizations of $Y|X$ at training inputs $x_i$. While this was sufficient to understand test-time performance in the older smoothing literature which historically focused on \textit{in-sample} (also known as \textit{fixed design}) prediction \citep{hastie1990GAM, rosset2019fixed}, \cite{curth2023u} argued that -- because the modern statistical learning literature is predominantly interested in \textit{generalization} to new (previously unseen) inputs (see e.g. \citet[Ch. 5.2]{goodfellow2016deep}; \citet[Ch. 4.1]{murphy2022probabilistic})  -- considering the behavior of smoothers at previously observed inputs is likely insufficient to understand their test-time performance. 

 For this reason, \cite{curth2023u} propose a generalized effective parameter measure  $p^{0}_{\hat{\mathbf{s}}}$ for a smoother $\hat{\mathbf{s}}$ which allows to measure the amount of smoothing conditional on a given set of (potentially new) inputs $\{x^0_j\}_{j \in \mathcal{I}_0}$:
\begin{equation}
\textstyle p^{0}_{\hat{\mathbf{s}}} \equiv  p(\mathcal{I}_{0}, \hat{\mathbf{s}}(\cdot))= \frac{n}{|\mathcal{I}_{0}|} \sum_{j \in \mathcal{I}_0} ||\hat{\mathbf{s}}(x^0_j)||^2
\end{equation}
This measure extends the \textit{variance-based} definition of effective parameters discussed in \citet[Ch. 3.5]{hastie1990GAM} to arbitrary sets of inputs. This definition was originally motivated by the fact that for a \textit{fixed and linear} smoother $\hat{\mathbf{s}}(\cdot)$ and for outcomes generated with homoskedastic variance $\sigma^2$, the variance of predictions is  $
    \text{Var}(\hat{f}(x_0)|\{x_i\}_{i \in \mathcal{I}_{\text{train}}}) = ||\hat{\mathbf{s}}(x_0)||^2 \sigma^2$.
Note that, as we discussed above, tree-based smoothers are neither fixed nor linear. As we further investigate in \cref{app:var}, this means that $p^{0}_{\hat{\mathbf{s}}}$ does not linearly predict variance in tree-based smoothers (yet, we find that it does predict their relative ranking in terms of variance well). However, we are mainly interested in $p^{0}_{\hat{\mathbf{s}}}$ for another reason: it allows to quantify how \textit{smooth} predictions are by providing a summary of the distribution of the weights $
\hat{\mathbf{s}}(x_0)$ -- intuitively, as we expand on below, the less uniform these weights are distributed over the training examples, the higher $p^{0}_{\hat{\mathbf{s}}}$ (and the less smooth the predictions).

\paragraph{Interpreting $ p^{0}_{\hat{\mathbf{s}}}$.}
For smoothers that output weighted averages with $\sum^n_{i=1}\hat{s}^i(x_0)=1$ and $0 \leq \hat{s}^i(x_0) \leq 1$, which holds for trees and forests when $\sum_b w_b=1$,  $ p^{0}_{\hat{\mathbf{s}}}$ is bounded as $1 \textstyle \leq p^{0}_{\hat{\mathbf{s}}} \leq n$, where the lower bound is attained by the training sample mean (i.e. a $n$-Nearest neighbor estimator, with completely uniform weights $\hat{s}^i(x_0)=\frac{1}{n}$ for all $i$) and the upper bound is attained by e.g. a $1$-Nearest neighbor estimator (or any other estimator that has $\hat{s}^j(x_0)=1$ for some $j$ and $\hat{s}^k(x_0)=0$ for all $k \neq j$). Thus, the higher $\textstyle p^{0}_{\hat{\mathbf{s}}}$, the less smooth the smoother when issuing predictions for a set of inputs $\mathcal{I}_0$.

Beyond relative scales, values of $p^{0}_{\hat{\mathbf{s}}}$ are hard to interpret on their own. However, we note that the magnitude of $p^{0}_{\hat{\mathbf{s}}}$ can be understood more intuitively by further considering an analogy with their behavior for nearest neighbor estimators. For a k-NN estimator, we have that $p^{0}_{\hat{\mathbf{s}}}=\frac{n}{k}$, which is why -- for smoothers satisfying $\sum^n_{i=1}\hat{s}^i(x_0)=1$ and $\hat{s}^i(x_0)\geq 0$ for all $i$ -- one could also interpret $\tilde{k}^0_{\hat{\mathbf{s}}}=\frac{n}{p^{0}_{\hat{\mathbf{s}}}}$ (with $1\leq \tilde{k}^0_{\mathbf{s}} \leq n$) as capturing something akin to the \textit{effective number of nearest neighbors} used by this adaptive nearest neighbor smoother. More precisely, $\tilde{k}^0_{\hat{\mathbf{s}}}$ corresponds to $k$ for which a k-NN estimator would attain the same number of effective parameters as $\hat{\mathbf{s}}(\cdot)$.

\subsection{Why train- and test-time smoothing behaviors of trees and forests are likely to differ}\label{sec:effp-diff} 

Finally, we build intuition for why the effective parameters used for prediction are likely to differ between (i) predictions issued for training and testing examples and (ii) trees and ensembles thereof, before we empirically investigate this behavior in the following sections. Why such differences in behavior could be expected is easiest to see for interpolating trees and forests, i.e. models trained to achieve perfect training performance, in regression problems -- thus we rely on this example below. 

This is because in regression with unique outcome realizations for every training sample, interpolation can only be achieved by trees grown to full depth -- i.e. trees for which every training example falls into its own leaf. That means that for inputs $x_i$ observed at training time the smoothing weights for any interpolating tree will be given by $\hat{\mathbf{s}}(x_i)=\mathbf{e}_i$, where  $\mathbf{e}_i$ is the unit vector with $1$ at position $i$, \textit{deterministically} (despite randomness in the realized tree structure). The same therefore holds for ensembles of such interpolating trees. Thus, for any interpolating regression tree and any interpolating forest, we will always have $\textstyle p^{train}_{\hat{\mathbf{s}}} = n$ on the training data. That is, both individual interpolating trees and interpolating ensembles will always behave like a 1-NN estimator \textit{on the training data}. 

When issuing predictions for a previously unseen test example $x_0$, on the other hand, the behavior of a single tree will likely \textit{not} be the same as that of a forest. As before, note that any individual tree will place $x_0$ into a leaf with a single training example so that $\hat{\mathbf{s}}(x_0)=\mathbf{e}_j$ for some training input $x_j$. Thus, an individual interpolating tree behaves like a $1-$NN estimator with $\textstyle p^{test}_{\hat{\mathbf{s}}} = n$ for unseen inputs too. However, an ensemble of interpolating trees will \textit{not} necessarily behave like a $1-$NN estimator on unseen examples: this is because the decision boundary is likely to be underdetermined at new $x_0$ and therefore different random trees may place $x_0$ into terminal leaves with different training examples. This is illustrated for a simple stylized example in \cref{fig:page1fig}. That is, for two interpolating trees we may have $\hat{\mathbf{s}}^{1, tree}(x_0)=\mathbf{e}_j$ and $\hat{\mathbf{s}}^{2, tree}(x_0)=\mathbf{e}_{j'}$ with $j \neq j'$. Then, while each individual tree behaves like a 1-NN estimator at $x_0$, their average in a 2-tree forest $\hat{s}^{forest}(x_0)=\frac{1}{2}(\hat{\mathbf{s}}^{1, tree}(x_0)+\hat{\mathbf{s}}^{2, tree}(x_0)) = \frac{1}{2}(\mathbf{e}_j+\mathbf{e}_{j'})$ will be a 2-NN estimator with $\textstyle p^{test}_{\hat{\mathbf{s}}} = \frac{n}{2}$. 

Therefore, interpolating forests should be expected to (i) be smoother when issuing predictions for testing examples than for training examples and (ii) be smoother than individual trees at previously unseen test inputs -- and this property is \textit{self-enforced} through the randomness in the tree-construction process.

\section{Revisiting, refining and reconciling recent explanations of Random Forest success}\label{sec:reconcile}
In this section, we use the above smoother perspective on trees and forests, and the effective parameter measure $\textstyle p^{0}_{\hat{\mathbf{s}}}$ in particular, to revisit, refine and reconcile \cite{wyner2017explaining} and \cite{mentch2020randomization}'s recent explanations for forest success. We begin in \cref{sec:wynersec} by revisiting \cite{wyner2017explaining}'s explanation by \textit{quantifying} their conjectured ``spiked-smooth'' behavior of interpolating forests: as alluded to in \cref{sec:effp-diff}, we now experimentally show that interpolating forests indeed use fewer effective parameters on previously unseen test examples than on training examples -- providing the first quantitative empirical evidence for ``spiked-smooth'' behavior in forests. We then also demonstrate that this behavior is \textit{not} actually unique to \textit{interpolating} forests. Next, we revisit the ``degrees of freedom'' explanation of \cite{mentch2020randomization}  in \cref{sec:mentchsec}. We show that their chosen measure for degrees of freedom cannot fully explain the success of forests relative to trees, but that replacing it by $\textstyle p^{0}_{\hat{\mathbf{s}}}$ discussed in the previous section can overcome this shortcoming. Finally, in \cref{sec:recsec}, we highlight that, while \cite{mentch2020randomization} consider their explanation to \textit{compete} with \cite{wyner2017explaining}'s, the latter explanation for forest success may actually be best understood as a \textit{special case} of \cite{mentch2020randomization}'s ``randomization as regularization'' argument. 

\paragraph{Experimental setup.} Throughout, to empirically illustrate our arguments, we use the ``MARSadd'' simulation adapted in \cite{mentch2020randomization} from \citet{friedman1991multivariate}, with \begin{equation}
Y=0.1e^{4X_1}+ \frac{4}{1+e^{-20(X_2-0.5)}}+3X_3+2X_4+X_5+\epsilon
\end{equation}
where errors are i.i.d. sampled as $\epsilon \sim \mathcal{N}(0, \sigma^2)$ and the $d\!\!=\!\!5$ features $\mathbf{X} \in [0, 1]^d$ are sampled independently from $Unif(0,1)$. We use $n_{train}\!\!=\!\!n_{test}\!\!=n\!\!=\!\!500$. We control the signal-to-noise ratio in the data-generating process by varying $\sigma$ but use $\sigma\!\!=\!\!1$ as a default unless otherwise indicated. All experiments are averaged over 10 replications with different random seeds and shaded areas in plots represent 95\% confidence intervals. In \cref{app:experiments}, we provide additional results that replicate our analyses on multiple real datasets from \citet{grinsztajn2022tree}, which include more samples, more features and classification outcomes. 

\subsection{Quantifying \cite{wyner2017explaining}'s conjectured ``spiked-smooth'' behavior of tree ensembles}\label{sec:wynersec}
\cite{wyner2017explaining} presented a novel perspective on the success of \textit{interpolating} tree-based ensembles. They conjecture that interpolating tree-based ensembles (i.e. ensembles composed of individual models that each fit the training data \textit{perfectly} without error) succeed because they are \textit{``spiked-smooth''} as they fit noise only very locally -- creating sharp regions around training examples (`spike') where one may wish to retain precise knowledge of the known label, while elsewhere in the input space the ensemble predictions are \textit{smoothed} out by averaging predictions across multiple models.
Below, we show that by interpreting tree ensembles as \textit{smoothers} and then measuring the effective amount of smoothing they do, we can \textit{quantify} the phenomenon conjectured by \cite{wyner2017explaining} and highlight that it is \textit{not} unique to interpolating models; to the best of our knowledge we are the first to do so.

\subsubsection{Quantifying the ``spiked-smooth'' behavior of interpolating tree ensembles}\label{sec:wynerexp}

\begin{figure}[!t]
    \centering
    \includegraphics[width=.99\textwidth]{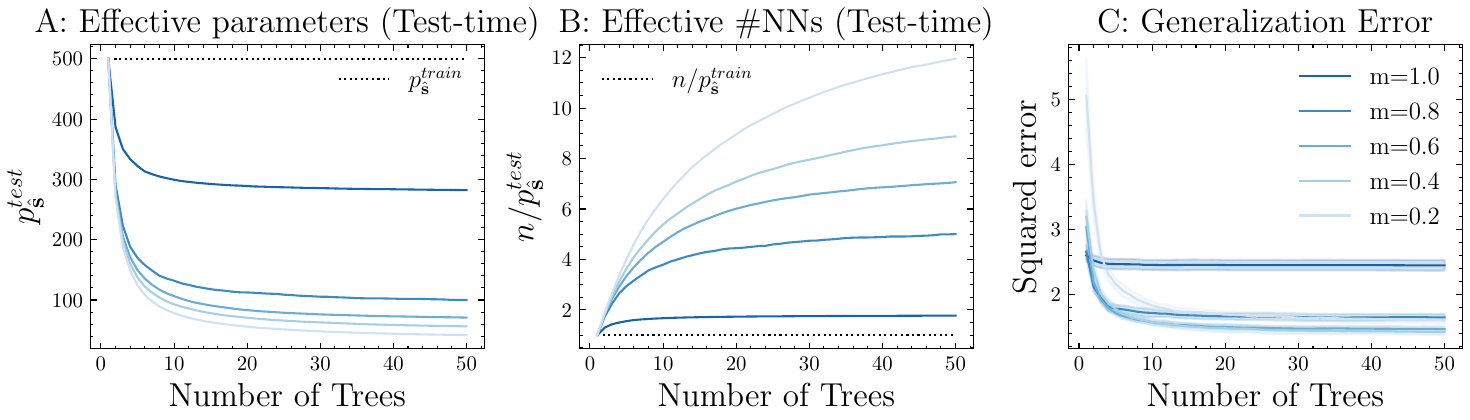}\vspace{-.3cm}
    \caption{\textbf{Understanding the performance of interpolating tree ensembles.} \small Effective parameters (left), effective number of nearest neighbors (middle) and generalization error (right)  by number of trees  for forests of full-depth trees trained without bootstrapping.}
    \label{fig:all-by-m}
\end{figure}

In \cref{fig:all-by-m}, we train a number of interpolating tree-ensembles (\textit{without} bootstrapping to enable interpolation of the complete training data by each tree) with different levels of randomness in their construction, and examine the behavior of the effective number of parameters as we increase the number of ensemble members. In the left panel of \cref{fig:all-by-m}, we find that all such interpolating forests indeed use the same number of effective parameters when issuing predictions for the \textit{training} examples \textit{regardless} of the number of trees and amount of randomness used (dotted black line). This is in line with the discussion in \cref{sec:effp-diff}: all interpolating forests behave like 1-NN estimators on the training data. The behavior is very different for \textit{unseen} testing examples (blue lines): the more randomness there is in tree construction, achieved by varying the proportion $m$ of features considered for every split, the more  $\textstyle p^{test}_{\hat{\mathbf{s}}}$ decreases as we increase the number of trees in the ensemble. These results thus clearly illustrate that more randomness and more trees result in smoother forest predictions. To investigate this further, we also plot the effective number of nearest neighbors $\tilde{k}^0_{\mathbf{s}}=\frac{n}{ p^{test}_{\hat{\mathbf{s}}}}$ implied by the different models in the middle panel of \cref{fig:all-by-m}, which indeed confirms that forests of interpolating trees, which themselves behave like 1-NN estimators, act like k-NN estimators with $k>1$ at test-time. We also observe that increasing randomness in the tree-building process (by decreasing $m$) corresponds to levels of smoothing consistent with k-NN estimators with increasing k. Overall, the gap between $p^{train}_{\hat{\mathbf{s}}}$ and $\textstyle p^{test}_{\hat{\mathbf{s}}}$  thus perfectly quantifies the spiked-smooth behavior conjectured by \cite{wyner2017explaining}: unlike individual trees, forests use \textit{less} effective parameters on testing than on training examples and are thus indeed more smooth in regions where no training point was observed previously.

\begin{wrapfigure}{r}{0.3\textwidth}
    \centering\vspace{-0.45cm}
\includegraphics[width=0.3\textwidth]{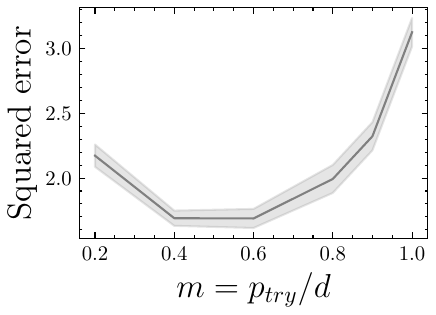}\vspace{-.4cm}
    \caption{\textbf{Generalization error by $m$ } \small for ensembles of 50 interpolating trees.}\label{fig:u-rand}\vspace{-.3cm}
\end{wrapfigure}Finally, in the right panel of \cref{fig:all-by-m}, we plot the generalization error of ensembles of different size for different levels of randomness in tree construction to relate the smoothing effects of ensemble size and randomness to the ensembles' prediction performance. On the one hand, we observe that the smoothing effects achieved through adding more trees never hurt: generalization error either drops or stays constant, depending on the original size of the ensemble. On the other hand, we observe that smoothing achieved by increasing randomness in tree construction, controlled through the proportion $m$ of features considered for each split, has a U-shaped effect on generalization error. This becomes easier to see in \cref{fig:u-rand} where we plot a cross-section of the right plot of \cref{fig:all-by-m}, considering only ensembles of size $B=50$ and instead place $m$ on the x-axis. We observe that introducing more randomness (by decreasing $m$) appears to improve generalization initially, until a turning point is reached when additional randomness hurts performance -- this is when tree construction becomes \textit{too} random. This behavior is analogous to the behavior of k-NN estimators with varying $k$ \citep[Ch. 2.3]{hastie2009elements}, and thus further cements the parallels between k-NNs and tree ensembles as \textit{adaptive} neighborhood smoothers (recall from \cref{fig:all-by-m} that setting $m$ larger implies larger $\tilde{k}^{test}_{\mathbf{s}}$).

\subsubsection{``Spiked-smooth'' behavior is \textit{not} unique to interpolating models}\label{sec:spike-smooth}

\begin{figure}[!t]
    \centering
    \includegraphics[width=.99\textwidth]{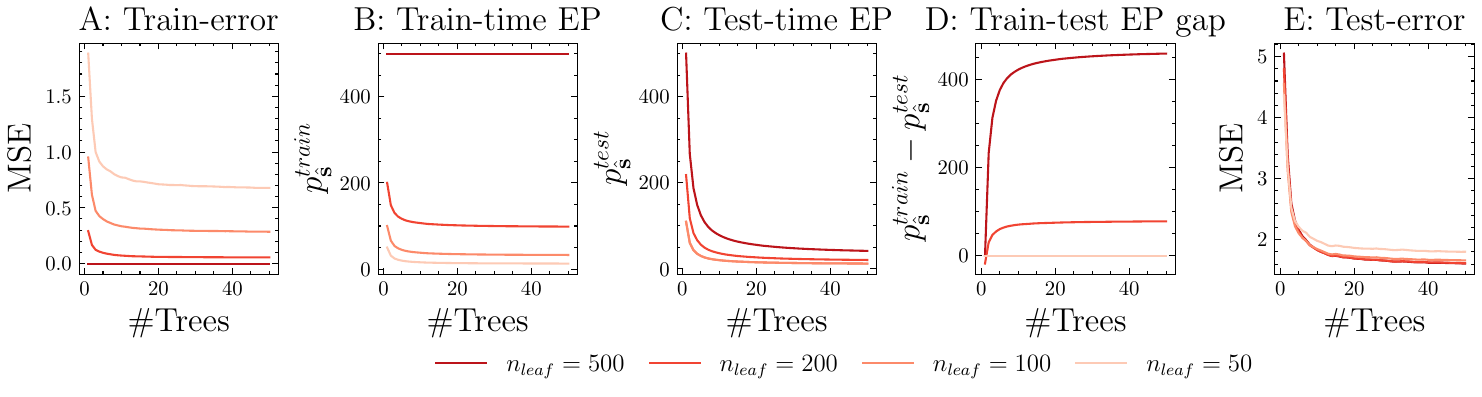}\vspace{-.3cm}
    \caption{\textbf{The smoothing effect of ensembling for trees of different depth.} \small Training error (A), train-time effective parameters (EPs, B), test-time effective parameters (C),  the difference (gap) between train-time and test-time effective parameters (D) and generalization error (E) by number of trees for forests of trees of different depths trained without bootstrap and with $m=\frac{1}{3}$.}
    \label{fig:train-test-gap}
\end{figure}

Equipped with a way of \textit{quantifying} `spiked-smooth' behavior -- i.e. models being smoother when making predictions for previously unseen testing points than for points seen during training --  we can now investigate whether this behavior is unique to models that fit the training data perfectly. To do so, we control the trees' ability to fit the training data by varying the maximum allowed number of leaves $n_{leaf}$ (while fixing $m=\frac{1}{3}$, as is usually recommended \cite[Ch. 15]{hastie2009elements}, and continuing to disable bootstrapping as above). In doing so in \cref{fig:train-test-gap}, we make a number of interesting observations.  In the first panel of \cref{fig:train-test-gap}, we confirm that reducing $n_{leaf}$ indeed limits the models' ability to fit the training data perfectly. In the second panel of \cref{fig:train-test-gap}, we observe that adding additional trees to a forest \textit{does} reduce the number of effective parameters used for \textit{training} examples if the individual models cannot interpolate (i.e. if $n_{leaf}<500$). This means that, in non-interpolating forests, training examples also  have more effective neighbors in forests than in individual trees -- this is different from the behavior of interpolating forests as studied in the preceding experiments. In the third panel of \cref{fig:train-test-gap}, we show that the test-time effective parameters also decrease as the number of trees in increase for all tree depths, as expected. In the fourth panel of \cref{fig:train-test-gap}, we now test whether there is ``spiked-smooth'' behavior -- using fewer effective parameters on unseen examples than on training examples, as measured by the effective parameter gap  $\textstyle p^{train}_{\hat{\mathbf{s}}} - \textstyle p^{test}_{\hat{\mathbf{s}}}$ -- in non-interpolating forests. Indeed, we find that this behavior is \textit{not} unique to interpolating models -- yet, it is \textit{much more pronounced} in interpolating trees as the train-test effective parameter gap shrinks rapidly as tree-depth decreases. We thus conjecture that spiked-smooth behavior appears whenever there is some degree of \textit{overfitting} to the training data. In the fifth panel of \cref{fig:train-test-gap}, we finally observe virtually no difference in the generalization errors of the ensembles with 100, 200 and 500 leaves, despite wide gaps between their training-error plotted in Panel A.

\begin{wrapfigure}{r}{0.45\textwidth}
    \centering \vspace{-.2cm}
\includegraphics[width=0.45\textwidth]{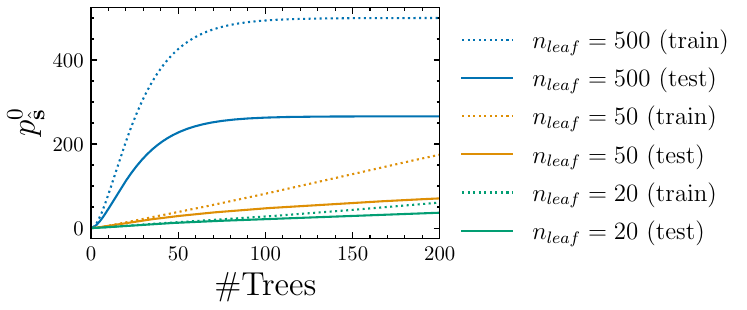}\vspace{-.4cm}
    \caption{\textbf{Train- and test-time effective parameters for boosted ensembles} \small using gradient boosting of trees of different depths with learning rate $\gamma=.05$.}\label{fig:boost}
\end{wrapfigure}\textbf{Boosted ensembles.} In \cref{fig:boost}, we demonstrate that -- as conjectured by \cite{wyner2017explaining} -- boosted ensembles, in this case constructed through gradient boosting,  also display a ``spiked-smooth'' property: regardless of the depth of the used trees, the effective parameters used for training and testing inputs begin to diverge as more ensemble members are added. Unlike the standard tree ensembles considered before, the number of effective parameters is \textit{increasing} in the number of trees. This is entirely expected due the way that gradient-boosted ensembles are constructed -- the  learning rate (here $\gamma=.05$) controls the contribution of individual trees and the ensemble weights (which do not add up to 1). Once more, we can also observe that ``spiked-smooth'' behavior is \textit{not} unique to ensembles constructed of trees that can interpolate.

\subsection{Where \cite{mentch2020randomization}'s ``degrees of freedom'' explanation falls short of explaining forest success (and how to fix it)}\label{sec:mentchsec}
Next, we revisit  \cite{mentch2020randomization}'s ``degrees of freedom''(DOF) explanation for  Random Forest success. In particular, they show that, in tree ensembles trained with bootstrapping, increasing randomness through $m$ has a regularizing effect as it reduces the DOF of the learned predictor $\hat{f}$ as quantified by $df(\hat{f}) = \frac{1}{\sigma^2} \sum_{i \in \mathcal{I}_{train}} Cov(\hat{y}_i, y_i)$ \citep{efron1986biased, tibshirani2015degrees}, which can be understood as measuring the dependence of train-time predictions $\hat{y}_i$ on \textit{their own} train-time labels $y_i$.\footnote{Like \cite{mentch2020randomization} we estimate the  $df(\hat{f})$ through Monte Carlo evaluation of the covariance formula. In particular, we resample $y_i$ and fit all models $50$ times to compute the covariance in each replication of the base experiment.}

\begin{figure}[!t]
    \centering
    \includegraphics[width=.99\textwidth]{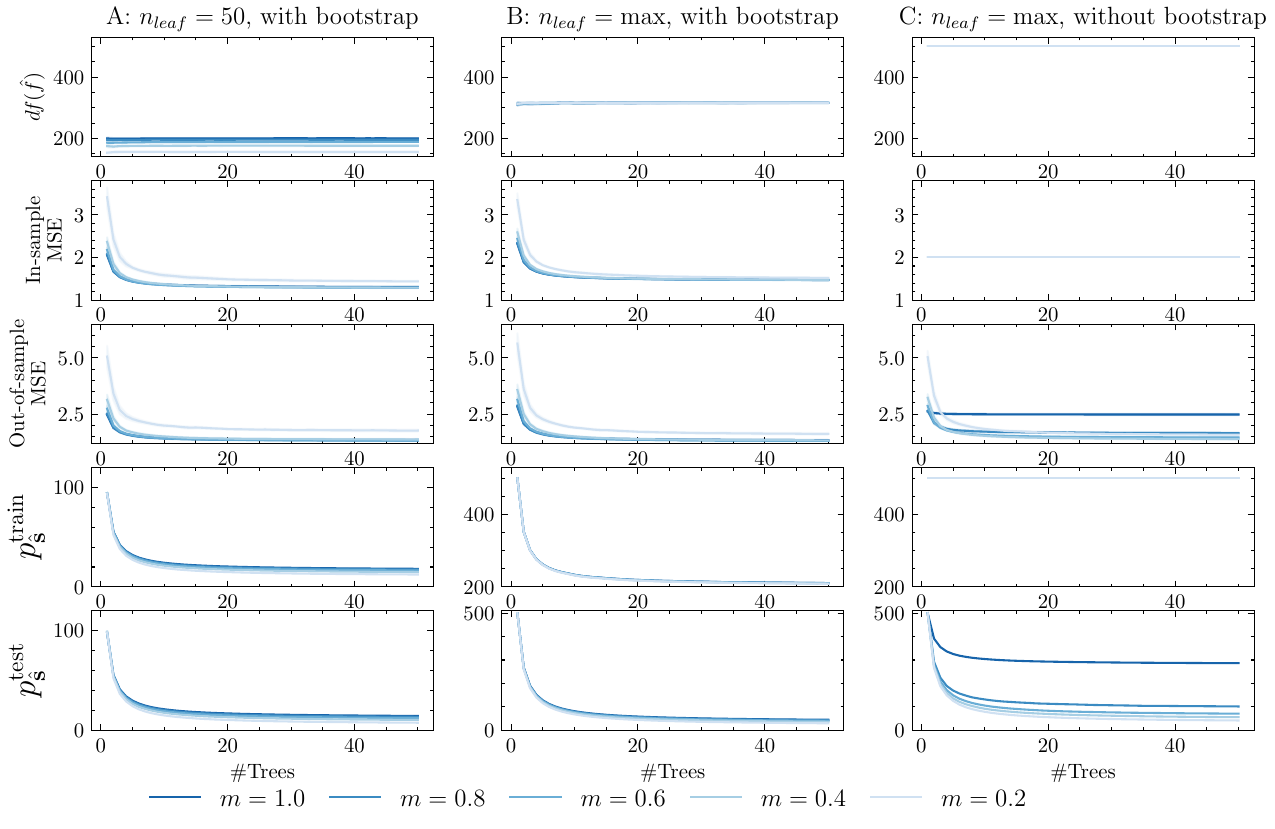}\vspace{-.3cm}
    \caption{\textbf{Degrees of freedom as measured by $df(\hat{f})$ do not always correlate well with model prediction performance (both in- and out-of-sample). Effective parameters as measured by $\textstyle p^{0}_{\hat{\mathbf{s}}}$ sometimes better reflect differences in model performance. } \small Plotting $df(\hat{f})$ (row 1), in-sample mean-squared error (MSE) (row 2), out-of-sample MSE (row 3), train-time effective parameters  $\textstyle p^{train}_{\hat{\mathbf{s}}}$ (row 4) and test-time effective parameters $\textstyle p^{test}_{\hat{\mathbf{s}}}$ (row 5)   for ensembles constructed with different levels of randomness (indicated by color) and different hyperparameters (different columns) against ensemble size on the x-axis. }
    \label{fig:mentch-main}
\end{figure}

Here, we show that measuring DOFs using the above covariance-based  definition $df(\hat{f})$ considered by \cite{mentch2020randomization} falls short of fully explaining random forest performance. This is highlighted by multiple empirical observations in \cref{fig:mentch-main}: First, we consider ensembles of trees with limited depth trained with bootstrapping as in \cite{mentch2020randomization} (column (A)). For fixed ensemble size, $ df(\hat{f})$ indeed differs across different levels of $m$ (row 1). For a given $m$, however, individual trees and ensembles composed of different numbers of trees all have \textit{the same} average DOF as measured by $df(\hat{f})$. Yet, these models clearly exhibit different test-time performances, both in terms of in-sample (or: fixed design) prediction error (where we resample new noisy outcomes $y'_i$ for training inputs $x_i$, row 2) and generalization error (which samples entirely new input-output pairs $(x'_j, y'_j)$ for testing, row 3).  Thus, while $df(\hat{f})$ can explain differences in performance across different $m$ for these ensembles, it falls short of explaining differences in test performance across different ensemble sizes. Second, we repeat the same experiment for ensembles of full-depth trees trained \textit{with} bootstrapping (column (B)). In this case, we observe that individual trees and ensembles composed of different numbers of trees and trained with different levels of randomness $m$ all have almost identical average DOF as measured by $ df(\hat{f})$ despite some differences in prediction performance both across $m$ and across varying ensemble sizes. Finally, we consider the behavior for ensembles of full-depth trees trained without bootstrapping (i.e. allowing interpolation as in \cref{sec:wynersec}) in column C. In this case, we observe that individual trees and ensembles composed of different numbers of trees and trained with different levels of randomness $m$ all have exactly \textit{the same} $ df(\hat{f})$. This correlates well with their \textit{in-sample} prediction error which is also constant (as is expected, as all interpolating models necessarily make the same prediction on training data),  yet these models clearly all exhibit different test-time performances in terms of generalization error. Thus, for ensembles trained without bootstrapping, $df(\hat{f})$ also falls short of explaining differences in generalization performance across \textit{both} different ensemble sizes \textit{and} different $m$.

To explain this possibly unexpected behavior of $df(\hat{f})$, we first note that $df(\hat{f})$, just like other classical measures of effective parameters, was developed in the so-called \textit{fixed design} setting, where it was assumed that the test inputs will be \textit{the same} as the training input points (see e.g. \citet[Ch. 3.5]{hastie1990GAM}). In this setting, correlation between train-time outcomes and predictions at training inputs can provide some indication of overfitting. However, when we are interested in \textit{generalization} to new input points, the train-time correlation captured by $df(\hat{f})$ gives us little information about a predictor's test-time behavior. This explains why $df(\hat{f})$ does not correlate well with generalization performance in the third row of \cref{fig:mentch-main}. Why then does $df(\hat{f})$ \textit{also} not correlate with the in-sample test error of ensembles of different sizes in scenarios (A) and (B)? We conjecture that this could be because the \textit{expected} dependence of a train-input prediction on its own train-time label is independent of the number of trees used (i.e. the expected value of $s^i(x_i)$, which determines $df(\hat{f})$, is constant across $B$) -- this is what $df(\hat{f})$ captures. Yet, there can be more smoothing across \textit{all other} training labels as $B$ grows (i.e. more uniform expected $s^j(x_i)$, $j\neq i$) due to the randomness induced by bootstrapping -- which may impact prediction performance. 

Fortunately, as we demonstrate in the bottom two rows of \cref{fig:mentch-main}, both shortcomings can be overcome by replacing $df(\hat{f})$ with \cite{curth2023u}'s $\textstyle p^{0}_{\hat{\mathbf{s}}}$. On the one hand, distinguishing the effective degrees of smoothing applied to train and test inputs allows to overcome the first issue. This is particularly visible in scenario (C) where $\textstyle p^{train}_{\hat{\mathbf{s}}}$ is constant as expected while $\textstyle p^{test}_{\hat{\mathbf{s}}}$ drops in ensemble size. On the other hand, the decrease of $\textstyle p^{train}_{\hat{\mathbf{s}}}$ in ensemble size in scenarios (A) and (B) indeed highlight that the second issue is in fact overcome by $\textstyle p^{train}_{\hat{\mathbf{s}}}$ as it measures smoothing across \textit{all} labels (instead of only dependence on their own label as $df(\hat{f})$ does).

\subsection{Resolving tensions between the ``spiked-smooth interpolation'' and ``randomness-as-regularization'' explanations of forest success}\label{sec:recsec}
In their paper, \cite{mentch2020randomization} consider \cite{wyner2017explaining}'s ``spiked-smooth interpolation'' explanation for random forest success insufficient for two main reasons: (i) if bootstrapping is used, as is the default for many forest implementations, trees cannot always interpolate the training data perfectly -- understanding their success as interpolating classifiers is thus inherently limited as an explanation for Random Forest success in practice. Further, (ii) \cite{wyner2017explaining} appear to ignore the role of additional randomness due to feature subsampling in forest success, as ``the explanation the authors provide for random forest success would seem to apply equally well to bagging''\cite[p.8]{mentch2020randomization}.  \cite{mentch2020randomization} then argue that Random Forests instead work specifically because of the regularization effect achieved by the randomness induced by the hyperparameter \textit{m}. Below, we wish to briefly highlight that once one takes into account the results presented in the preceding sections, the two explanations do not actually stand in conflict with each other -- rather, we argue that \cite{wyner2017explaining}'s explanation should be understood as a special case of \cite{mentch2020randomization}'s more general one.

\begin{figure}
\centering
\begin{minipage}{.45\textwidth}
  \centering
  \includegraphics[width=.99\linewidth]{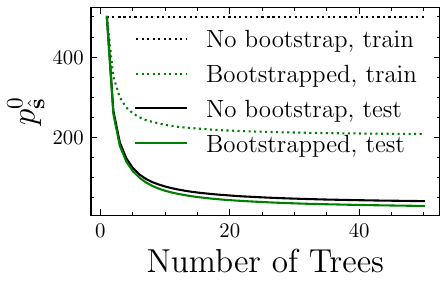}\vspace{-.3cm}
  \caption{\textbf{The effect of bootstrapping on effective parameters} \small of trees grown to full depth ($m=\frac{1}{3}$).}
\label{fig:booteffect}
\end{minipage}%
\hfill
\begin{minipage}{.45\textwidth}
  \centering
  \includegraphics[width=.99\linewidth]{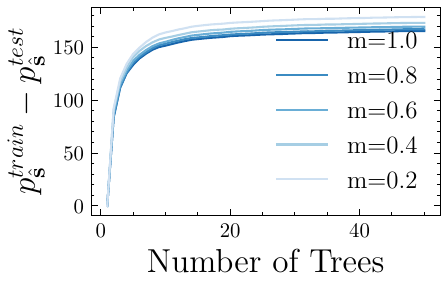}\vspace{-.3cm}
  \caption{\textbf{Train-test effective parameter gap for different $m$ }\small of bootstrapped tree ensembles grown to full-depth.}
    \label{fig:m-effect}
\end{minipage}
\end{figure}

 In particular, with regards to point (i), recall that we showed in \cref{sec:spike-smooth} that the performance-driving \textit{spiked-smooth} behavior conjectured by \cite{wyner2017explaining} is not unique to interpolating models -- so (i) is not necessarily a limiting factor of their ``spiked-smooth'' argument for forest success\footnote{Of course, \cite{wyner2017explaining} did not make this point themselves, and we agree with \citet{mentch2020randomization} and \cite{zhou2023trees} that it is not necessarily \textit{interpolation itself that causes} this favorable spiked-smooth behavior -- as is a key point of \cite{wyner2017explaining}'s argument.}. Indeed, we highlight in \cref{fig:booteffect} that tree-ensembles are spiked-smooth even when bootstrapping is applied (and interpolation can hence not be achieved) as there is a gap between train- and test-time effective parameters also in this case. Further, we discover in \cref{fig:booteffect} that the addition of bootstrapping itself has a smoothing effect on the \textit{train-time} predictions, similar to restriction of tree-depth in \cref{fig:train-test-gap}.

With regards to point (ii), using our approach of \textit{quantifying} spiked-smooth behavior through train-test differences in $\textstyle p^{0}_{\hat{\mathbf{s}}}$, we can now show that the spiked-smooth behavior of forests is in fact \textit{a direct result of the randomization induced by the tree-building process} (as without randomness, all trees would be identical and no smoothing on the test-examples could occur). Indeed, as we show in \cref{fig:m-effect}, spiked-smooth behavior is directly related to $m$ (even in bootstrapped ensembles).

Thus, spiked-smooth behavior \textit{is a consequence} of regularization induced by randomization, and \cite{wyner2017explaining}'s explanation is better understood as \textit{a special case} of the ``randomization as regularization''  perspective of \cite{mentch2020randomization}. Indeed, both explanations can be best understood as relying on the \textit{self-regularizing} properties of tree ensembles achieved through randomization in their construction. \textit{Both} randomness in split selection and randomness induced through bootstrapping have a \textit{smoothing} effect on forest predictions, which can differ between train and test inputs.

\section{Why do forests generalize better than trees? Rethinking the bias-variance effects of ensembling}\label{sec:bias-var}
In the previous section, we demonstrated that ensembling can result in forest predictions that are \textit{smoother} than the predictions of individual trees. In this section, we will investigate \textit{why} this improves the test error of forests relative to trees. Today, it appears to be widely accepted that improvements of forests over individual trees are a result of \textit{variance reduction only} (\citealp[Ch. 15.4.2]{hastie2009elements}, \citealp{mentch2020randomization, ghosal2020boosting}). Yet, \citet[p. 29]{breiman2001random} conjectured in the conclusion of his seminal Random Forest paper:  ``Their accuracy indicates that they act to reduce bias. The mechanism for this is not obvious'' and \citet[p.4]{dietterich2002ensemble}'s similarly hinted at a bias-reducing effect of ensembling more generally. Next, we will therefore re-open the discussion into whether and why tree ensembling could also be understood to also have a \textit{bias-reducing effect} (\cref{sec:theorybiasvar}). We will argue that, while trees and forests indeed have the same (systematic) bias in mean (which is what the term ``bias'' generally refers to in statistics \citep[Ch. 1]{lehmann2006theory}), they do differ in what functions they can represent and can thus differ in another kind of bias (hinted at in \citet{dietterich2002ensemble}). We will then experimentally demonstrate that the variance-reduction and bias-reduction intuitions are useful in different contexts: in \cref{sec:snr} we consider the effect of ensembling at different levels of signal-to-noise ratio in outcomes on both in- and out-of-sample prediction error. We observe that, while in-sample prediction improves through (bagged) ensembling only when noise is high, when generalization to new inputs is of interest, forests can outperform trees even in settings with noiseless outcomes. In \cref{sec:bias}, we then empirically show that, for previously unseen input points, differences in what trees and forests can represent drives differences in error. 

\subsection{Understanding mechanisms for bias- and variance-reduction in tree ensembles}\label{sec:theorybiasvar}
In this section, we re-investigate in what ways ensembling of trees could affect ``bias'' and ``variance'' of learned predictors. We believe that both the terms ``bias-reduction'' and ``variance-reduction'' are somewhat overloaded across different literatures (statistics and computer science in particular), and will therefore attempt to be precise in distinguishing between different mechanisms that may sometimes implicitly be covered by these terms.

\paragraph{Preliminaries.} Throughout, we consider the problem of estimating the expected outcome $\mu(x_0)=\mathbb{E}[Y|X=x_0]$ for some input point $x_0$ using estimators $\hat{\mu}(x_0)\equiv \hat{\mu}(x_0; Z, \Omega)$, which are random variables because they depend on the random training sample $Z$ and may also use some additional randomness summarized in $\Omega$. We assume we are interested in finding estimators  $\hat{\mu}(x_0)$ that minimize the $\text{EMSE}(\hat{\mu}(x_0))= \mathbb{E}_{Z, \Omega}[(\mu(x_0)-\hat{\mu}(x_0))^2]$. This is equivalent to minimizing the prediction error for $Y|X=x_0$ using the squared loss, whose value differs from the EMSE only in the additional constant term $\text{Var}(Y|X=x_0)$. 

\subsubsection{The statistics perspective on bias and variance in ensembles}\label{sec:statsview}
We begin by examining the \textit{statistical} perspective on the bias-variance effects of ensembling as presented in e.g. \citet[Ch. 15.4]{hastie2009elements}. In terms of bias, this literature is concerned with whether estimators $\hat{\mu}(x_0)$ make \textit{systematic errors} in estimating the true $\mu(x_0)=\mathbb{E}[Y|X=x_0]$, i.e. whether they \textit{systematically} over- or underestimates the truth \citep[Ch. 1]{lehmann2006theory}. This leads to a definition of bias, which we will refer to as as \textit{statistical bias (StatBias)}, given by $\text{StatBias}(\hat{\mu}(x_0)) = \mu(x_0) - \mathbb{E}_{Z, \Omega}[\hat{\mu}(x_0)]$. It is easy to see that any two estimators with the same expected value $\mathbb{E}_{Z, \Omega}[\hat{\mu}(x)]$ will always incur identical StatBias. \citet[Ch. 15.4.2]{hastie2009elements} show that individual trees and forests constructed using i.i.d. draws of $\Theta$ indeed have the same expected value $\bar{T}(x_0)$:
\begin{equation*}
    \mathbb{E}_{Z, \mathbf{\Theta}}[{F}_{B, \mathbf{\Theta}}(x_0)] = \mathbb{E}_{Z, \mathbf{\Theta}}\left[ \frac{1}{B}\sum^B_{b=1} T(x_0; \Theta^b)\right] = \mathbb{E}_{Z, {\Theta}}[T(x_0; \Theta)] = \mathbb{E}_{Z}[\mathbb{E}_{\Theta|Z}[T(x_0; \Theta)]]=\bar{T}(x_0)
\end{equation*}
which immediately implies that forests and trees will incur the \textit{same} StatBias. Because $\text{EMSE}(\hat{\mu}(x_0))=\text{StatBias}(\hat{\mu}(x_0))^2 + \text{Var}_{Z, \Omega}(\hat{\mu}(x_0))$ by the well-known bias-variance decomposition \citep[Ch. 7.3]{hastie2009elements},  from this statistical perspective \textit{all} improvements of forests relative to trees \textit{must} thus be due to variance reduction alone. 

Therefore, consider next the variance of forest predictions, which \citet[Ch. 15.4.1]{hastie2009elements} show can be written as: 
\begin{equation}\label{eq:var-decomp}
    \text{Var}_{Z, \mathbf{\Theta}}({F}_{B, \mathbf{\Theta}}(x_0)) = \underbrace{\text{Var}_Z(\mathbb{E}_{\mathbf{\Theta}|Z}[{F}_{B, \mathbf{\Theta}}(x_0)]}_{\text{(i) SampVar$(\hat{\mu}(x_0))$}} + \underbrace{\mathbb{E}_Z[\text{Var}_{\mathbf{\Theta}|Z}({F}_{B, \mathbf{\Theta}}(x_0))]}_{\text{(ii) WithinZVar$(\hat{\mu}(x_0))$}}
\end{equation}
This in turn highlights that increasing $B$ can reduce the total variance of predictions in two ways: It may reduce (i) the sampling variation ($\text{SampVar}(\hat{\mu}(x_0))$) -- i.e. the impact of noise in the training data-generating process -- and it can reduce (ii) the within-Z variance ($\text{WithinZVar}(\hat{\mu}(x_0))$) -- i.e. the variability in which predictor is \textit{realized} in practice given any particular training dataset. The first mechanism would be important in settings with low signal-to-noise ratio (SNR), which is where $\text{Var}(Y|X=x_0)$ is high, because limited SampVar would imply limited overfitting to noise in outcomes. This is precisely in line with the explanation for forest success given in \cite{mentch2020randomization}. As we discuss below, the second term implicitly \textit{absorbs multiple other mechanisms} for forest success -- and we will argue that some of them may indeed actually be better understood as impacting \textit{a form of bias of the model class}. 

\subsubsection{The computer science perspective on bias and variance in ensembles}\label{sec:cs-view}
Next, we revisit an older computer science perspective on the potential sources of gain of ensembling (in general) as discussed in \citet{dietterich2000ensemble, dietterich2002ensemble}.  Importantly, this perspective disregards sampling variability due to $Z$. Instead, \citet{dietterich2002ensemble} (informally) discusses three reasons why averaging over multiple learnable functions learned from a given training sample (in our context, this would be trees constructed using different realizations of $\Theta$) may lead to improved prediction performance. \citet{dietterich2002ensemble}'s ``statistical reason'' is that many learnable functions are indistinguishable on the training data but may have different generalization performance.  Their ``computational reason'' is that model search is often in some way only \textit{locally} optimal due to greedy optimization procedures. Finally,  their ``representational reason'' is that ensembling may be able to expand the space of representable functions relative to the functions that individual members can represent. Within the statistical perspective from \cref{sec:statsview}, all three reasons would impact the within-Z variance of \cref{eq:var-decomp}, yet \citet{dietterich2002ensemble} actually considers only the first two reasons variance-reducing mechanisms and the final reason \textit{a bias-reducing} mechanism. Below, we will therefore formalize \citet{dietterich2002ensemble}'s arguments to resolve the ostensible tension between the two perspectives. 

To do so, let $\mathcal{H}_Z(\hat{\mu})=\{\hat{\mu}(\cdot; Z, \Omega): P(\Omega|Z)>0\}$ denote the class of functions that can be learned by an algorithm from training data $Z$ (e.g. $\mathcal{H}_{Z}(T_\Theta)=\{T_\Theta(\cdot): P(\Theta|Z)>0)\}$ for trees). Then, for a test point $x_0$ let $\hat{\mu}^*_Z(x_0)$ denote the value of the \textit{best} learnable function in $\mathcal{H}(Z)$, i.e. $\hat{\mu}^*_Z(x_0) \in \arg \min_{h \in \mathcal{H}_Z(\hat{\mu})} (h(x_0)-\mu(x_0))^2$. In this context, we would argue that \citet{dietterich2002ensemble}'s first two reasons correspond to reductions in what we will refer to as \textit{model variability} (ModVar): $\text{ModVar}_Z(\hat{\mu}(x_0)) = \mathbb{E}_{\Omega|Z}[(\hat{\mu}(x_0) - \hat{\mu}^*_Z(x_0))^2]$. This measures the risk that our error is dominated by ``bad draws'' of predictors $\hat{\mu}(x_0)$ relative to the best attainable predictor $\hat{\mu}^*_Z(x_0)$. Conversely, we would argue that the final reason of \citet{dietterich2002ensemble}, i.e. reductions in representational bias of the model class,  concerns the performance of the best attainable $\hat{\mu}^*_Z(x_0)$ relative to the ground truth $\mu(x_0)$. We will therefore formalize this \textit{representation bias} (RepBias) as $\text{RepBias}_Z(\hat{\mu}(x_0)) = (\hat{\mu}^*_Z(x_0) - \mu(x_0))^2$.

A natural next question is whether we can relate 
$\text{ModVar}_Z(\hat{\mu}(x_0))$ and $\text{RepBias}_Z(\hat{\mu}(x_0))$ to $\text{EMSE}(\hat{\mu}(x_0))$ in a similar way to the classical bias-variance decomposition in statistics. Indeed, because $(a+b)^2\leq 2(a^2 + b^2)$, we can relate the quantities as
\begin{equation}
    \text{EMSE}(\hat{\mu}(x_0))
    \leq 2 \cdot \mathbb{E}_{Z}\left[\underbrace{(\mu(x_0)- \hat{\mu}^*_Z(x_0))^2}_{\text{RepBias}_Z(\hat{\mu}(x_0))} + \underbrace{\mathbb{E}_{\Omega|Z}[(\hat{\mu}^*_Z(x_0) - \hat{\mu}(x_0))^2]}_{\text{ModVar}_Z(\hat{\mu}(x_0))}\right] 
\end{equation}
Further, note that whenever $\hat{\mu}^*_Z(x_0)=\mathbb{E}_{\Omega|Z}[\hat{\mu}(x_0)]$, the more direct relationship \\$\text{EMSE}(\hat{\mu}(x_0))=\mathbb{E}_{Z}[\text{RepBias}_Z(\hat{\mu}(x_0))+\text{ModVar}_Z(\hat{\mu}(x_0))]$ holds. 

\subsubsection{What is a meaningful notion of bias to understand the performance tree ensembles?}
The above discussion highlights that the distinction between the two perspectives mainly lies in whether the EMSE is expanded around the expected predictor $\mathbb{E}_{Z, \Omega}[\hat{\mu}(x_0)]$ (\cref{sec:statsview}) or the best in-class predictor $\hat{\mu}^*_Z(x_0)$ (\cref{sec:cs-view}). This begs the question: Is one more meaningful than the other in the context of trees? First, note that for a single tree and given training data, the expected predictor $\mathbb{E}_{\Omega|Z}[\hat{\mu}(x_0)]$ is likely to differ from the best in-class predictor $\hat{\mu}^*_Z(x_0)$. To see this in a simple example, consider again the stylized example of interpolating trees in \cref{fig:page1fig}. In this case, there are only two interpolating trees (displayed in Panels A and B).  Assuming each split is chosen with equal probability, the expected predictor $\bar{T}(\cdot)$ has the form of the ensemble depicted in Panel C. It is obvious that this expected predictor $\bar{T}(\cdot)$ \textit{cannot} be realized by growing individual trees to full depth and hence cannot be equal to the best in-class predictor in general. Second, this also highlights another peculiarity of trees: the expected predictor $\bar{T}(\cdot)$ is \textit{not} even necessarily a member of the hypothesis class $\mathcal{H}_{Z}(T_\Theta)$. That is, the class of trees \textit{cannot} necessarily represent \textit{its own expected predictor}. 

This demonstrates that, in the context of tree ensembles, the statistical notion of bias does not capture exactly what one might intuitively consider the bias of a learning algorithm  -- instead, it only carries information about whether the potentially unrealizable $\bar{T}(x_0)$ systematically over- or underestimates the true $\mu(x_0)$. We therefore argue that distinguishing trees and forests by their representation bias can be more meaningful. Why are trees and forests likely to differ in their RepBias? It is easy to see that the hypothesis class $\mathcal{H}_Z(F_{B, \mathbf{\Theta}})$ defined by forests can be much richer than the class of individual trees as it is given by:
\begin{equation*}
    \mathcal{H}_Z(F_{B, \mathbf{\Theta}}) = \left\{F_{B, \mathbf{\Theta}}(\cdot) = \sum_{b} w_b T_{\Theta^b}(\cdot) : \sum_b w_b = 1, w_b \in \left\{\frac{0}{B}, \frac{1}{B}, \ldots, \frac{B}{B}\right\}, T_{\Theta^b}(\cdot) \in \mathcal{H}_Z(T_{\Theta})\right\}
\end{equation*}
\begin{wrapfigure}{r}{0.45\textwidth}
    \centering \vspace{-.2cm}
\includegraphics[width=0.45\textwidth]{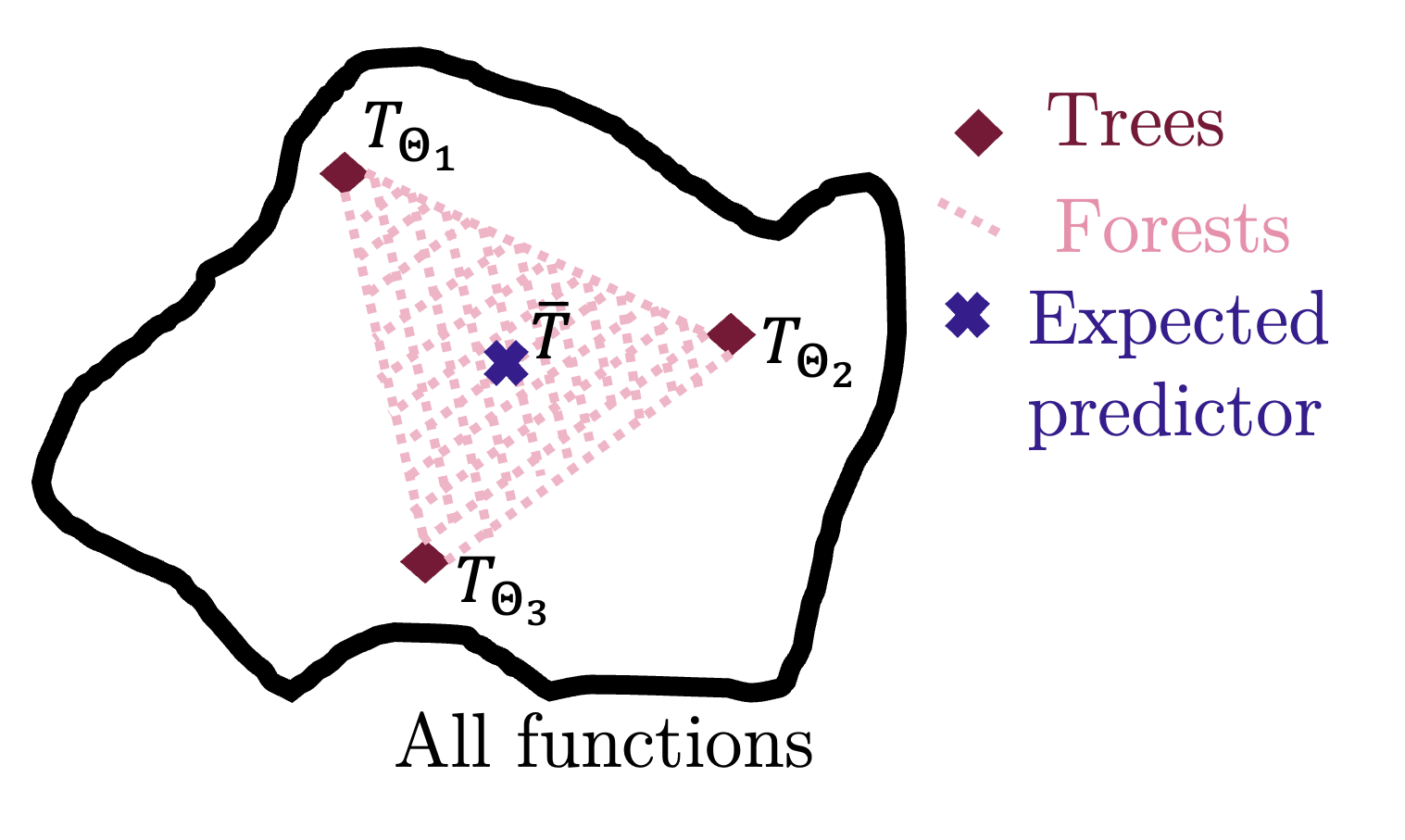}\vspace{-.4cm}
    \caption{\textbf{Illustration comparing the hypothesis classes spanned by trees and forests.}}\label{fig:hypclass}
\end{wrapfigure}As further illustrated in \cref{fig:hypclass} forests can thus be conceptualized as a grid on a simplex with vertices defined by individual trees. $\mathcal{H}_Z(F_{B, \mathbf{\Theta}})$ therefore also becomes richer as $B$ increases. This may be most intuitive when revisiting the example of interpolating trees and forests: when considering their implied smoother weights, individual interpolating trees will always have $\hat{\mathbf{s}}(x_0)=\mathbf{e}_j$ for some training index $j$ with a single non-zero entry, while interpolating forests can have $\hat{\mathbf{s}}(x_0)$ with up to $B$ non-zero entries (depending on $P(\mathbf{\Theta}|Z)$), and can thus represent different predictions than individual trees can. Note also that, as $B$ increases, forests become more likely to be able to represent the expected predictor $\bar{T}(x_0)$

As $\mathcal{H}_Z(T_{\Theta}) \subset \mathcal{H}_Z(F_{B, \mathbf{\Theta}})$, we will always have $\text{RepBias}_Z(F_{B, \mathbf{\Theta}}(x_0)) \leq \text{RepBias}_Z(T_{\Theta}(x_0))$. Conversely, the expected relative behavior of their model variability terms is less clear as there are two competing mechanisms at play: on the one hand, a richer class is likely to also contain \textit{worse} hypotheses (increasing ModVar), while on the other hand ensembling could lead to bad solutions being chosen with less probability (decreasing ModVar).

\subsubsection{Potential mechanisms for forest success}\label{sec:hypothesis-mechs}
Given the discussion above, we conjecture that there are \textit{three separate mechanisms} through which forests improve upon individual trees by smoothing the predictions:
\begin{enumerate}[noitemsep]
    \item Reduction of the sampling variation (SampVar) in cases where SNR is low
    \item Reduction of  model variability (ModVar) as the likelihood of outputting a bad predictor decreases
    \item Reduction of the representation bias (RepBias) by enriching the class of learnable functions 
\end{enumerate}

\subsection{Empirically testing the different mechanisms for forest success}\label{sec:emp-bias-var}
Next, we will empirically test whether and when the different mechanisms identified above drive the relative performance of trees and forests. We continue to use the MARSAdd simulation discussed in \cref{sec:reconcile}. We begin in \cref{sec:snr} by investigating the impact of outcome noise, and then close in \cref{sec:bias} by investigating how model variability and representation bias play a role when predicting outcomes at previously unseen test inputs. 

\subsubsection{Understanding the impact of outcome noise on forest performance}\label{sec:snr}
First, we investigate how varying the signal-to-noise ratio (SNR) in outcomes through the error variance $\sigma^2$ changes the relative performance of forests and trees.  As we discussed in the preceding section, the SNR impacts the sampling variation ($\text{SampVar}(\hat{\mu}(x_0))$): when there is little signal in outcomes, predictors that are overfitted to single labels may experience high error. In this context,   \cite{mentch2020randomization} concluded that random forests should be particularly successful in settings with low SNR (high $\sigma^2$) due to the regularizing effects of randomness in the tree-building process. To revisit this question, we investigate the in- and out-of-sample prediction performance of trees and forests grown to full-depth with and without bootstrapping. Recall that, as discussed in \cref{sec:recsec}, when in-sample prediction is of interest, only bootstrapped ensembles will experience smoothing through ensembling (as without bootstrapping all trees interpolate the training data).

\begin{figure}[!t]
    \centering
    \includegraphics[width=.99\textwidth]{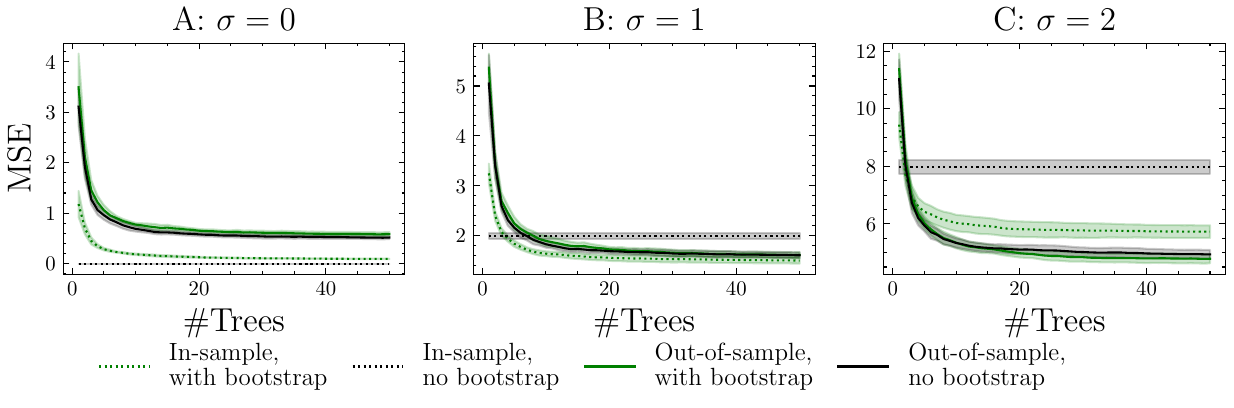}
    \caption{\textbf{Performance of tree ensembles at different outcome noise levels $\sigma$. }\\ \small In-sample prediction error (dotted lines) and out-of-sample prediction error (solid lines) by ensemble size of full-depth trees trained with (green) and without (black) bootstrapping (with $m=\frac{1}{3}$), for different outcome noise levels $\sigma$. The smoothing effects achieved by bagging improve in-sample prediction error whenever there is noise in outcomes (B \& C), but worsen performance relative to interpolating trees and forests when there is no noise (A). When considering out-of-sample prediction error, the smoothing effect achieved by ensembling trees with and without bootstrapping always improves upon individual trees regardless of noise level.} 
    \label{fig:SNR}
\end{figure}

In \cref{fig:SNR}, we study both the in-sample prediction error and the generalization error of tree ensembles consisting of full-depth trees, trained with and without bootstrapping, for different outcome noise levels $\sigma$. We observe that the behavior of the in-sample prediction error is as expected by \cite{mentch2020randomization}: the lower the SNR (i.e. higher $\sigma^2$), the more the smoothing effect of ensembling bagged trees improves upon individual trees in-sample. Conversely, when there is no noise at all ($\sigma=0$), the smoothing induced by bootstrapping actually \textit{hurts} in-sample performance: because full-depth trees trained to interpolate the training data can attain zero in-sample test-error when there is no noise in outcomes, the bagged ensemble performs worse than individual interpolating trees (or ensembles thereof, which also interpolate). This observation is in line with classical bias-variance tradeoff arguments \citep[Ch. 3.3]{hastie1990GAM}: smoothing on training inputs can decrease variance in predictions if there is noise in outcomes, but will come at the cost of increased bias because simply using $y_i$ to predict future outcomes at training inputs $x_i$ has \textit{no bias}. 

The behavior of generalization error, i.e. prediction error on previously unseen inputs, differs from the in-sample prediction error in a striking manner: when given previously unseen test examples, Panel (A) of \cref{fig:SNR} highlights that forest predictions can improve upon individual trees even \textit{in the absence of any label noise} ($\sigma=0$)! Thus, as conjectured in the previous section, there must be another mechanism by which ensembling improves tree predictions than just correcting for low SNR -- which we will investigate next. 

\subsubsection{Understanding the impact of model variability and representation bias}\label{sec:bias}
In the previous section, we observed that the relative performance of trees and forests differs depending on whether in-sample and out-of-sample prediction error is of interest. In the context of full-depth interpolating trees (grown without bootstrapping), this is a natural finding: note that the realizable predictions at \textit{training inputs} $x_i$ are the same for individual interpolating trees and forests thereof. That is, as noted in \cref{sec:effp-diff}, both act as 1-NN estimators $\hat{\mu}(x_i)=y_i$ and thus satisfy $\mathbb{E}_{\Omega|Z}[\hat{\mu}(x_i)]=\mu(x_i)=\mu^*_Z(x_i)$. That is, interpolating trees and forests have \textit{no bias of any form} when issuing predictions for training inputs. Any (in-sample) prediction error at training points of such interpolating models is thus due to noise in observed outcomes alone. This is different for new test points $x_0$ not seen during training: there is no perfect match $x_i$ in the training sample for which $\mu(x_0)=\mu(x_i)$ holds necessarily, thus we may additionally expect some form of bias to appear. Given this conceptual difference between in- and out-of-sample prediction, we anticipate that the difference between tree and forest predictions (and hence possible differences in bias due to differences in realizable predictions) should increase when we \textit{move further away} from the training inputs. 

Next, we will therefore test this intuition by, instead of sampling test points completely at random as before, now also controlling the average dissimilarity between train and test points. Given the $n \times d$ train input matrix $\mathbf{X}_{train}$, we now create test inputs as $\mathbf{X}_{test}=\mathbf{X}_{train}+\mathbf{X}_{offset}$ by adding a $n \times d$ random offset matrix with individual entries sampled from a uniform $U(-\delta, +\delta)$ distribution. The larger we choose $\delta$, the larger the expected distance between testing points and their nearest training neighbor. 
\begin{figure}[t]
    \centering
    \includegraphics[width=.8\textwidth]{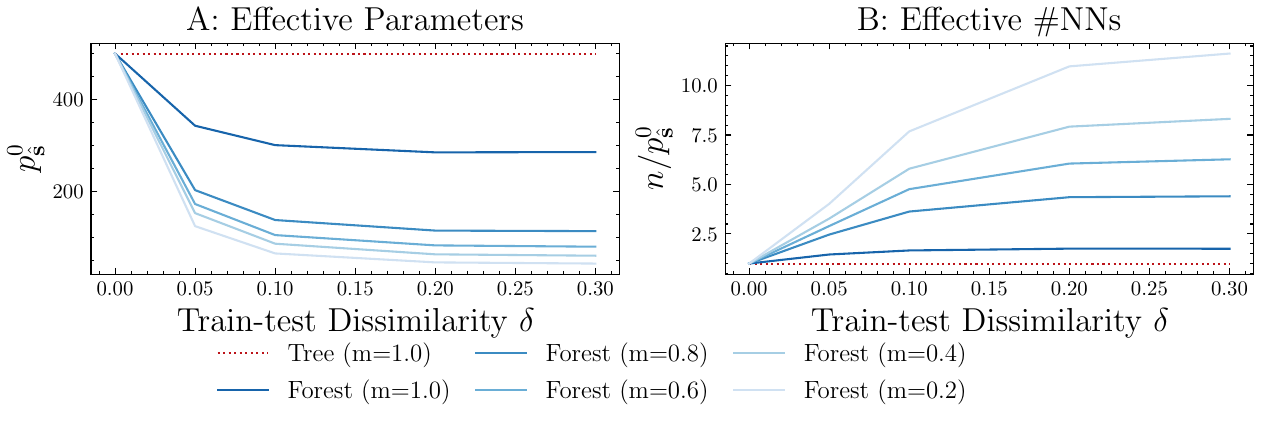}
    \caption{\textbf{The test-time smoothing behavior of forests relative to trees \textit{increases} in the dissimilarity $\delta$ between training and testing inputs.} \small Test-time effective parameters (left) and effective number of nearest neighbors (right) of interpolating tree-ensembles ($B=50$, without bootstrapping) by train-test dissimilarity, for $\sigma=0$.}\label{fig:effp-dist}
\end{figure}

\paragraph{How does train-test input dissimilarity affect the effective degree of smoothing?}
 In \cref{fig:effp-dist}, we use this setup to first examine how the test-time effective parameters and implied effective nearest neighbors evolve as we increase the train-test input dissimilarity through $\delta$ in a noiseless setting ($\sigma=0$). As before, single interpolating trees can only act as 1-NN estimators -- their $\textstyle p^{test}_{\hat{\mathbf{s}}}$ is constant at $n$ regardless of $\delta$. This is different for the interpolating forests with different degrees of randomness $m$: these, once more, display a self-regularizing property. As before, at any given $\delta$, more randomness implies smoother predictions. Controlling the train-test dissimilarity through $\delta$ additionally leads to an interesting new observation: we observe that for given $m$ the effective degree of smoothing \textit{increases} as the testing data moves \textit{further} from the training data ($\delta$ increases). Importantly, note that as we vary $\delta$, only the \textit{testing data} changes; the training data stays the same and there is thus \textit{no retraining}. That is, the observed change in smoothing behavior at different levels of $\delta$ thus appears automatically at test-time (without any change in the trained predictor). 

\begin{figure}[!t]
    \centering
    \includegraphics[width=.99\textwidth]{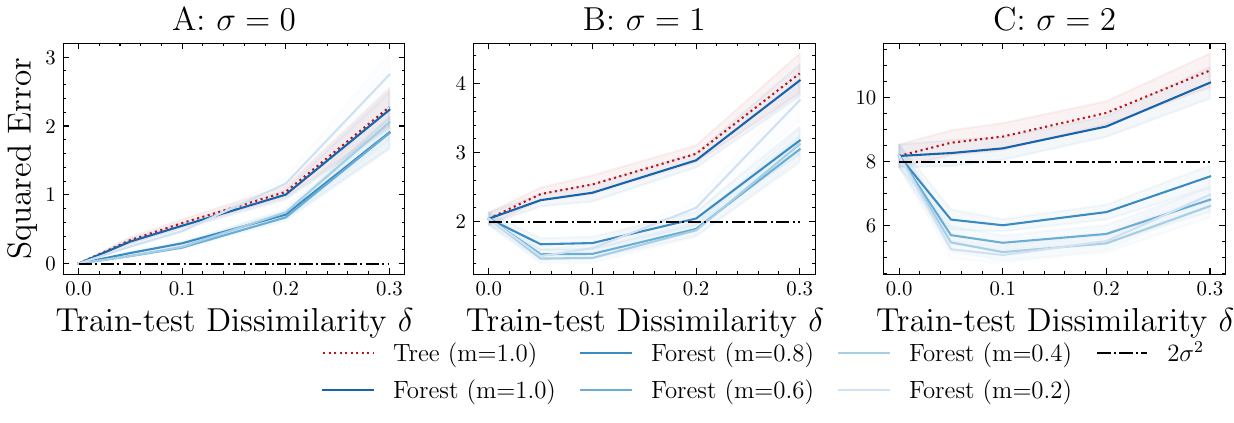}\vspace{-.3cm}
    \caption{\textbf{Generalization performance by train-test dissimilarity $\delta$ at different levels of noise $\sigma$} \small for interpolating trees and ensembles thereof ($B=50$, without bootstrapping). The gap in performance between an individual tree and forests with injected randomness ($m<1$) grows as we move away from the training data due to differences in predictions that trees and forests can realize, even for noiseless outcomes. When there is noise in outcomes, there is additionally an outcome-variance-reduction effect at play that widens the performance gap.}
    \label{fig:mse-by-dist}
\end{figure}

\paragraph{How does train-test input dissimilarity affect generalization performance?}
Next, we examine the resultant generalization behavior of forests relative to trees across varying $\delta$ and $\sigma$. In the left panel of \cref{fig:mse-by-dist}, we begin by examining the relative performance of individual trees and forests in a completely noise-free setting. When train- and test points coincide perfectly, trees and forests perform identically and make no errors, as is expected when $\sigma=0$. As the test points move further from the training points we observe that a gap between tree- and forest performance emerges for intermediate levels of $\delta$: as there is no label-noise in this setting, this is entirely due to the differences in the predictors that forests and trees realize. In particular, it seems that predicting by smoothing over \textit{more} training instances as forests do (cf. \cref{fig:effp-dist}) helps to generalize predictions to previously unobserved inputs in this setup. As before this is most beneficial for \textit{intermediate} values of the proportions $m$ of features considered at each split, while performance deteriorates for its extreme values. When the train-test dissimilarity controlled by $\delta$ becomes very large, the gap in performance between trees and forests becomes smaller as it becomes harder to generalize from the training data to the testing examples \textit{at all}.  Once we let $\sigma>0$, as in the middle and right panels of \cref{fig:mse-by-dist}, we observe that there is \textit{additionally} an outcome-variance-reduction effect at play: this is because, when training ($x^{train}_i$) and testing input  ($x^{test}_i$) are identical, using the 1-NN $y^{train}_i$ to predict $y^{test}_i$ will incur expected error $2\sigma^2$ -- thus, by smoothing over multiple instances, forests can improve upon trees when $\delta>0$ not only due to a difference in what they can represent but also due to a reduction in the variance of the predictions due to label noise.

\begin{figure}[!t]
    \centering
    \includegraphics[width=.99\textwidth]{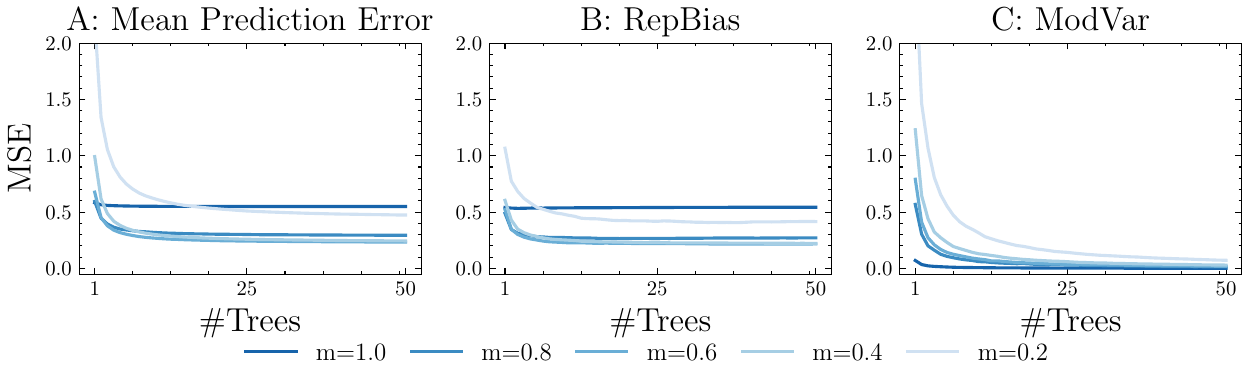}\vspace{-.3cm}
    \caption{\textbf{The effect of ensemble size on mean prediction error (A), representation bias (RepBias, B) and model variability (ModVar, C)} \small for interpolating trees. Evaluated in a setting without label noise ($\sigma=0$) and with intermediate train-test dissimilarity ($\delta=0.1$) by randomly re-initializing the same model 50 times and picking $\hat{\mu}^*_Z(\cdot)$ with oracle knowledge of the test outcomes. Ensembling improves upon individual trees constructed with the same level of randomness $m$ by decreasing \textit{both} the best-case prediction error \textit{and} model variability. When comparing individual tree constructed without additional randomness ($m=1$) to large forests with injected randomness $m<1$, we observe that reductions in RepBias achieved by ensembling appear to drive their superior performance.}
    \label{fig:error-decomp}
\end{figure}

\paragraph{Are differences in generalization error driven by representation bias or model variability?}
Finally, we can now examine more closely to what extent the observed difference in generalization behavior in the absence of label noise is indeed due to the last two mechanisms conjectured in \cref{sec:hypothesis-mechs}: does ensembling work because it reduces both the model variability (ModVar) \textit{and} the representation bias (RepBias)? To test this in \cref{fig:error-decomp}, we train 50 versions of each model (changing only random seeds used in model construction) on each fixed train-test set (we use the noiseless setting with $\sigma=0$ and an intermediate $\delta=0.1$). Among these 50 models, we measure performance on the test set and identify the best-performing model as $\hat{\mu}^*_Z(\cdot)$. (Note that identifying $\hat{\mu}^*_Z(\cdot)$  requires making use of oracle knowledge of the test outcomes). Its prediction performance $MSE(\hat{\mu}^*_Z(\cdot))$ is a proxy for RepBias (Panel B), while the average MSE between predictions of this best model and all 50 models is used to assess ModVar (Panel C). 

We make a number of interesting observations. First, in Panel B of \cref{fig:error-decomp}, we indeed find evidence that ensembling reduces RepBias: we observe that when there is some injected randomness in tree construction ($m < 1$), averaging multiple trees \textit{reduces} the best-case error substantially. This indicates that the smoother functions that the forests can learn indeed enrich the hypothesis space in a way that reduces RepBias relative to the true underlying outcome-generating process. \cref{fig:error-decomp} indicates that the best functions in this case are learned for intermediate values of randomness $m$, which corresponds to intermediate levels of smoothing (cf. \cref{fig:effp-dist}). Second, in Panel C of \cref{fig:error-decomp}, we also find evidence that ensembling additionally reduces the effect of ModVar: adding additional trees rapidly decreases the difference between individual randomly realized models and the best model at that ensemble size. As could be expected, we observe that the more randomness in tree construction there is, the more pronounced the ModVar reduction effect. Comparing Panels A, B and C, it becomes clear that it is the reduction in RepBias that drives the difference in prediction performance of forests trained with injected randomness ($m<1$) relative to a single tree grown without injected randomness ($m=1$). In \cref{app:experiments}, we demonstrate that these findings replicate also when using non-interpolating trees and on real-world datasets.

Summarizing the findings in this section, considering especially \cref{fig:mse-by-dist} and \cref{fig:error-decomp}, we conclude that when prediction performance \textit{at new inputs} is of interest, our results indicate that forests can generalize better than trees \textit{both} because they reduce the impact of label noise \textit{and} because they can realize a wider range of predictions.

%overfitting on yourself vs overfitting on some example: is there a difference conceptually?

%\section{what do htf say?}
%`` Hence the improvements in prediction obtained by bag- ging or random forests are solely a result of variance reduction.'' - is this exactly true? can't a forest represent something a tree cannot?
%\subsection{Connections to diversity?}

\section{Conclusion}
In this paper, we showcased how interpreting trees and ensembles thereof as \textit{adaptive smoothers} can provide new insight into the mechanisms by which Random Forests achieve their excellent performance in practice. In particular, we highlighted that this perspective allows us to \textit{measure} the effective amount of smoothing implied by such predictors -- which, in turn, allows us to quantify automatic regularizing effects of ensembling and increased randomness in the tree-building process. We also demonstrated that the level of smoothing can differ greatly between inputs observed at training time and completely new test-time inputs, reflected also in stark differences in the behavior of in-sample prediction error and generalization error of tree ensembles. Finally, we highlighted that forests can improve upon trees even in outcome noise free settings because they are able to realize different predictions, and used this observation to challenge the prevailing wisdom that ensembling should be understood as a tool for variance reduction alone. To the contrary, we showed that forests improve upon trees through \textit{multiple distinct} mechanisms that are usually implicitly entangled: they reduce the effect of noise in outcomes,  reduce the variability in realized predictors \textit{and} reduce potential bias by enriching the class of functions that can be represented. 

\paragraph{Practical takeaways.} Throughout our experiments, we observed that forests outperform individual trees because averaging predictors constructed with some inherent randomness can have a \textit{smoothing} effect on predictions. While this is beneficial in moderation, we observed that \textit{too much} randomness can also harm prediction performance. Further, we found that different ways of inducing randomness can have somewhat different effects: inducing randomness through bootstrapping has a much larger effect on in-sample than out-of-sample prediction performance, because it seems to mainly impact the ability to overfit to labels \textit{at previously observed inputs}. Randomness in feature selection, especially when the training data can be fit perfectly, is much more important for out-of-sample prediction performance. Finally, we observed that any smoothing induced by randomized ensembling is useful for in-sample prediction error only when the signal-to-noise ratio is low, but can improve out-of-sample prediction error even in the absence of any label noise -- because it allows to realize a wider range of predictions than individual models. Whether or not test inputs will likely differ from inputs already observed at training is thus an important consideration in practice.

\bibliography{references}
\newpage
\appendix
\section*{Appendix}
This appendix is structured as follows: In \cref{app:var}, we examine the relationship between $p^{0}_{\hat{\mathbf{s}}}$ and the variance of tree-based adaptive smoothers. In \cref{app:boost}, we show how to derive the smoothing weights implied by gradient boosted tree ensembles. Finally, in \cref{app:experiments}, we present additional experiments and additional information on the experimental setup.

\section{Examining the relationship between effective parameters $p^{0}_{\hat{\mathbf{s}}}$ and the variance of tree-based adaptive smoothers}\label{app:var}

In this section, we examine the relationship betweeen $p^{0}_{\hat{\mathbf{s}}}$ and predictive variance in tree-based smoothers. As we discussed in \cref{sec:smooth-back}, $p^{0}_{\hat{\mathbf{s}}}$ was originally derived in a setting where $\hat{\mathbf{s}}$ is linear and fixed conditional on training-inputs $\{x_i\}_{i \in \mathcal{I}_{\text{train}}}$ and therefore determines variance in predictions as $\text{Var}(\hat{f}(x_0)|\{x_i\}_{i \in \mathcal{I}_{\text{train}}}) =  ||\hat{\mathbf{s}}(x_0)||^2 \sigma^2$. In trees and forests, $\hat{\mathbf{s}}$ is neither linear nor fixed, thus we expect that $p^{0}_{\hat{\mathbf{s}}}$ will \textit{underestimate} the expected variance in predictions. Below, we demonstrate experimentally that while $p^{0}_{\hat{\mathbf{s}}}$ does allow to \textit{rank} different smoothers by their expected predictive variance, $p^{0}_{\hat{\mathbf{s}}}$ indeed underestimates the magnitude of variance.

\begin{figure}[t]
    \centering
    \includegraphics[width=.99\textwidth]{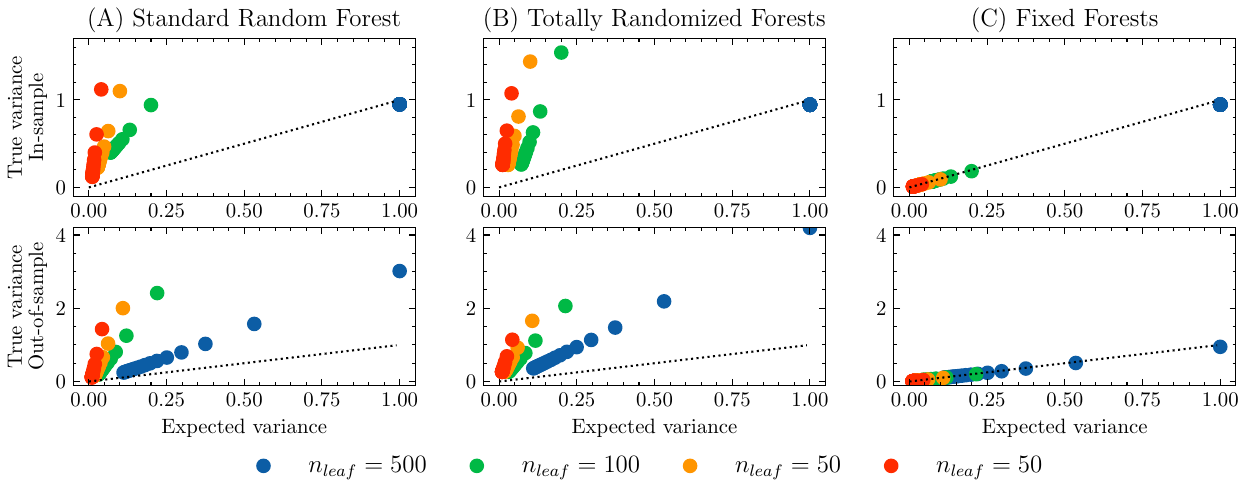}
    \caption{\textbf{True predictive variance versus expected variance according to $||\hat{\mathbf{s}}(x_0)||^2 \sigma^2$, in-sample (top) and out-of-sample (bottom)} for forests implemented in different ways (columns) with different depths (colors).}
    \label{fig:predictive-variance}
\end{figure}

In \cref{fig:predictive-variance}, we examine the relationship between the average predictive variance (both in- (top) and out-of-sample (bottom)) and $||\hat{\mathbf{s}}(x_0)||^2 \sigma^2$ in the MARSadd simulation with $\sigma^2=1$. We compute predictive variances by, for fixed training \textit{input} realizations, resampling training outcomes 20 times (and then refitting models); we then report averages of this over 10 replications with new random samples of inputs. As the standard random forests in column (A), we use forests with $m=\frac{1}{3}$ and no bootstrapping composed of trees of different depths (indicated by color), where the different points of the same color result from ensembles of different sizes (recall from \cref{sec:spike-smooth} that larger ensemble size $B$ implies lower $||\hat{\mathbf{s}}(x_0)||^2$). To disentangle the effects of adaptivity and randomization on predictive variance, we then consider multiple variants of standard forest implementations. To create non-adaptive yet random forests, we implement a version of totally randomized forests \citep{geurts2006extremely} by randomly shuffling the training labels for creation of the tree structures that determine $\hat{\mathbf{s}}(x_0)$ (but update the leaf predictions using the actual observed labels  $\mathbf{y}_{train}$ ) in column (B) of \cref{fig:predictive-variance}.  To create forests that are both non-adaptive and fixed in column (C), we create tree structures $\hat{\mathbf{s}}(x_0)$ based on a random permutation of training labels and fix those structures across all resamples of outcomes, and then update their predictions for each random realization of training labels only using the resampled $\mathbf{y}_{train}$ in the leaves.

In doing so, we make a number of interesting observations in \cref{fig:predictive-variance}. First, in panel (A), we observe that for standard random forests using $||\hat{\mathbf{s}}(x_0)||^2 \sigma^2$ indeed \textit{substantially} underestimates the predictive variance both in- and out-of-sample (with the exception of interpolating trees in-sample, which -- as expected -- simply have the variance $\sigma^2$ as they predict their own label). While smoother tree-ensembles of the same depth are \textit{ranked} correctly in terms of their predictive variance, $||\hat{\mathbf{s}}(x_0)||^2\sigma^2$ underestimates the total predictive variance. There is one major difference when comparing the in- and out-of-sample variance results: out-of-sample, deeper trees imply higher (true) predictive variance -- as one may intuitively expect. In-sample, however, we observe \textit{lower} true variance for deeper trees.

Next, we will investigate \textit{why} $|\hat{\mathbf{s}}(x_0)||^2\sigma^2$ underestimates predictive variance -- is it because trees are adaptive or because they are not fixed? -- which will also give insight into the latter observation. In column (B) we observe that by considering non-adaptive trees  -- by using totally randomized trees which can only \textit{increase} the randomness in the smoother weight construction -- leads to even more pronounced underestimation of the actual predictive variance. When using fixed trees in column (C)  we observe that  $||\hat{\mathbf{s}}(x_0)||^2\sigma^2$ \textit{does} predict actual predictive variance very well. This together indicates that the major source of the underestimated variance in panel (A) lies in the unaccounted variance in smoother weights. In fact, comparing columns (A) and (B) highlights that adaptivity actually \textit{counteracts} some of the effects of randomness in tree construction on the true predictive variance -- it reduces it by making some tree structures more likely than others. This also highlights that we likely observe lower true variance for deeper trees because these are \textit{more} adaptive to a training examples' own label and the total variance is thus less affected by variability in the choice of other contributors within their leaf.

\section{Boosted trees as smoothers}\label{app:boost}
In this section we show how boosted ensembles -- created using gradient boosting of regression trees -- can be interpreted as smoothers and present their associated smoothing weights $\hat{\mathbf{s}}(x_0)$. To do so, we begin by recalling the gradient boosting algorithm \citep{friedman2001greedy} (adapted from \citet[Ch. 10.10]{hastie2009elements}): 
\newpage
\textbf{Gradient boosted trees,  with learning rate $\eta$ and the squared loss as $\ell(\cdot, \cdot)$}

\begin{enumerate}
    \item Initialize $f_0(x)=0$
    \item For $p \in \{1, \ldots, P^{boost}\}$
    \begin{enumerate}[(a)] 
        \item For $i \in \{1, \ldots, n\}$ compute 
        \begin{equation}
            g_{i, p} = - \left[\frac{\partial \ell(y_i, f(x_i))}{\partial f(x_i)}\right]_{f=f_{p-1}}
        \end{equation}
        \item Fit a regression tree to $\{(x_i,  g_{i, p})\}^n_{i=1}$, giving leaves $L^{j}_{\Theta_p}$ for $j=1, \ldots, J_p$
        \item Compute optimal predictions for each leaf $j \in \{1, \ldots, J_p\}$:
        \begin{equation}\label{eq:boostpred}
            \gamma_{jp} = \arg \min_{\gamma \in \mathbb{R}} \sum_{i: l_{\Theta_p}(x_i) = L^{j}_{\Theta_p}} \ell(y_i, f_{p-1}(x_i) + \gamma) = \frac{1}{n_{l^{j}_{\Theta_p}}}  \sum_{i: l_{\Theta_p}(x_i) = L^{j}_{\Theta_p}} (y_i - f_{p-1}(x_i))
        \end{equation}
        \item Denote by $\tilde{T}(x_0, \Theta^p)=\sum^{J_p}_{j=1} \bm{1}\{l_{\Theta_p}(x_0) = L^{j}_{\Theta_p}\} \gamma_{jp}$ the predictions of the tree built in this fashion
        \item Set $f_p(x_0)=f_{p-1}(x_0) + \eta \tilde{T}(x_0, \Theta^p)$
    \end{enumerate}
    \item Output $f(x_0)=f_{P^{boost}}(x_0)$
\end{enumerate}

Appendix C.3.3 of \citet{curth2023u} then shows that smoothing weights of a boosted ensemble of $p$ gradient boosted regression trees can be constructed recursively as:

\begin{equation}
\hat{\mathbf{s}}^{boost, p}(\cdot) = \hat{\mathbf{s}}^{boost, p-1}(\cdot) + \eta \left(\hat{\mathbf{s}}^{tree, \tilde{f}_{p}}(\cdot) - \hat{\mathbf{s}}^{corr, f_{p}}(\cdot) \right)
\end{equation}

where $\hat{\mathbf{s}}^{tree, \tilde{f}_{p}}(\cdot)$ is the standard smoothing weights associated with the $p^{th}$ tree structure $\tilde{T}(x_0, \Theta^p)$ as given by \cref{eq:s_tree} in the main text and $\hat{\mathbf{s}}^{corr, f_{p}}(\cdot)$ is a correction term accounting for residualization in boosting as:
\begin{equation}
  \textstyle  \hat{\mathbf{s}}^{corr, f_{p}}(x_0) = \mathbf{e}_{l_{\Theta_p}(x_0)} \mathbf{\hat{R}}^{p} 
\end{equation}
 using the $1 \times J_p$ indicator vector $\mathbf{e}_{{l_{\Theta_p}(x_0)}}=(\bm{1}\{{l_{\Theta_p}(x_0)} = L^{1}_{\Theta_p}\}, \ldots, \bm{1}\{{l_{\Theta_p}(x_0)} = L^{J_p}_{\Theta_p}\})$ and where $\mathbf{\hat{R}}^{p}$ is the $J_p \times n$ leaf-residual correction matrix with $j-$th row given by
\begin{equation}
 \mathbf{\hat{R}}^{p} = \frac{1}{n_{l_{jp}}} \sum_{l_{\Theta_p}(x_i) = L^{j}_{\Theta_p}} \hat{\mathbf{s}}^{boost, p-1}(x_i)
\end{equation}

\section{Additional experiments}\label{app:experiments}

\subsection{Additional experiments on real world datasets}
In this section, we replicate some of the analyses in the main text on real world datasets. To do so, we use datasets from the tabular benchmark of \cite{grinsztajn2022tree}. We pick 4 datasets, as described in more detail below, that differ in their characteristics: we use two regression datasets (MiamiHousing2016, Superconduct)  and two binary classification datasets (Bioresponse, California), and for each use one with relatively few and one with relatively many features. In each experiment, we subsample all datasets to use $n_{train}=n_{test}=2000$ samples (we make an exception for the \textit{Bioresponse} dataset which only has 3434 samples in total, there we use $n_{train}=2000$ and $n_{test}=1400$). As in the main text we repeat this process 10 times for each dataset and report average metrics and 95\% confidence intervals based on 2 standard errors of the mean. 

\subsubsection{Regression datasets}
In this section, we use two real regression datasets to replicate some of the analyses from the main text. As a dataset with relatively few features, we use the MiamiHousing2016 dataset (\url{https://www.openml.org/d/44147}, $n_{total}=13,932$), in which the task is to predict (log) sales prices of single-family homes sold in Miami in 2016 from 13 features capturing house and location characteristics. As a second dataset, we use superconduct (\url{https://www.openml.org/d/44148}, $n_{total}=21263$), \cite{grinsztajn2022tree}'s regression dataset with most features, in which the task is to predict the critical temperature of superconductors using 79 features capturing their properties. 

\begin{figure}[b]
    \centering
    \includegraphics[width=.99\textwidth]{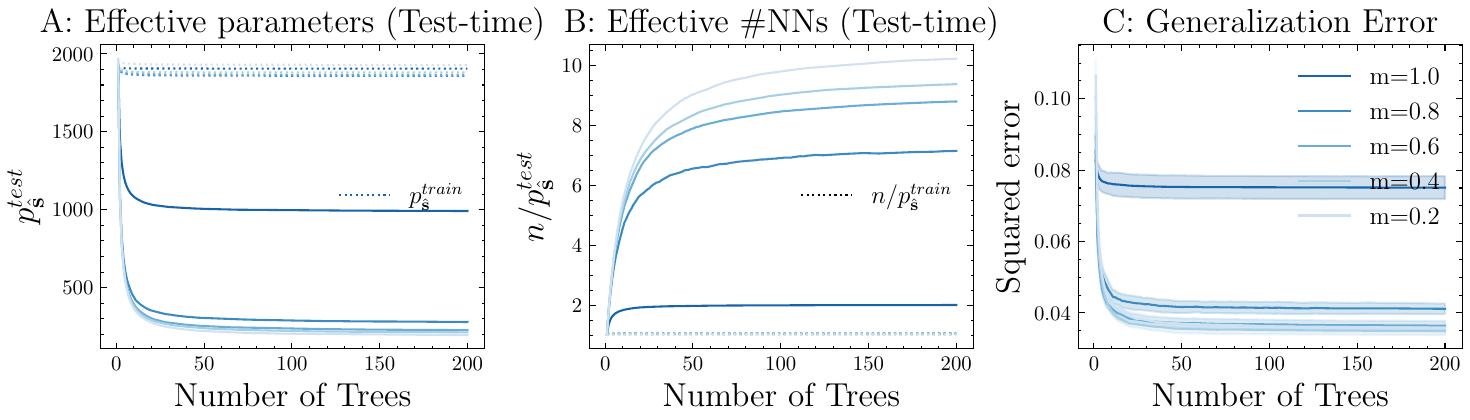}\vspace{-.3cm}
\caption{\textbf{Interpolating ensembles on the MiamiHousing2016 dataset (Regression, $d=13$).} }
    \label{fig:wyner-44147}
\end{figure}

\begin{figure}
    \centering
\includegraphics[width=.99\textwidth]{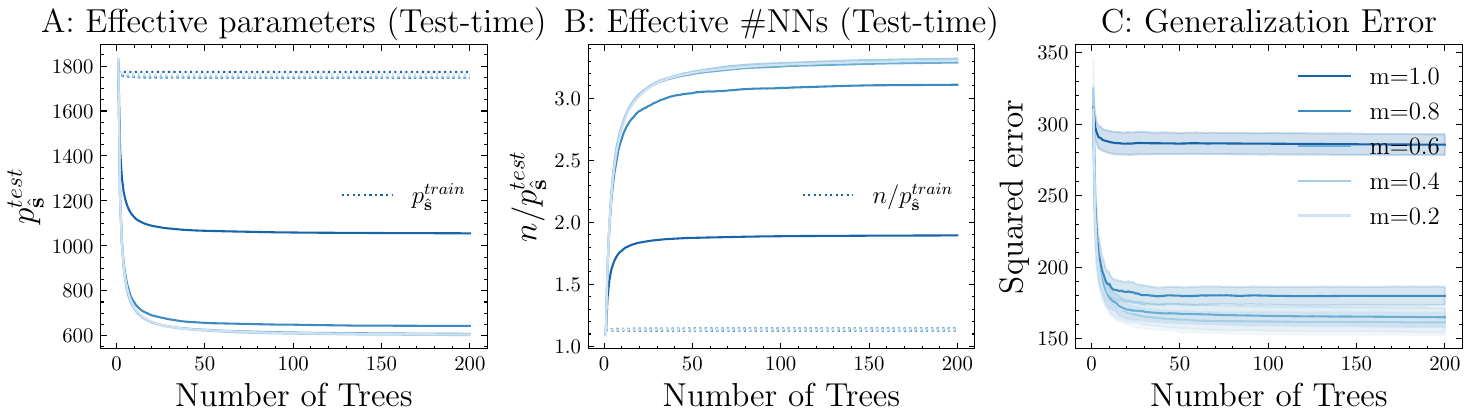}\vspace{-.3cm}
\caption{\textbf{Interpolating ensembles on the Superconduct dataset (Regression, $d=79$).} }
    \label{fig:wyner-44148}
\end{figure}

In \cref{fig:wyner-44147} and \cref{fig:wyner-44148}, we begin by replicating the experiment from \cref{sec:wynerexp} using forests of interpolating trees to verify that spiked-smooth behavior also occurs on real-world datasets. Indeed, we find that ensembling of interpolating trees has the expected smoothing effect also on the test examples in this real data, and this also improves their generalization error. We observe two main differences relative to the results on the simulated data in \cref{fig:all-by-m}: First, the train-time effective parameters $\textstyle p^{train}_{\hat{\mathbf{s}}}$ lie slightly below $n_{train}=2000$. This is because there are multiple examples with the same input values in this dataset, thus not all leaves contain a single training examples even in full-depth trees. Second, unlike in the simulated setting, we observe no deterioration of performance when setting $m$ very small.

\begin{figure}[h]
    \centering
\includegraphics[width=.99\textwidth]{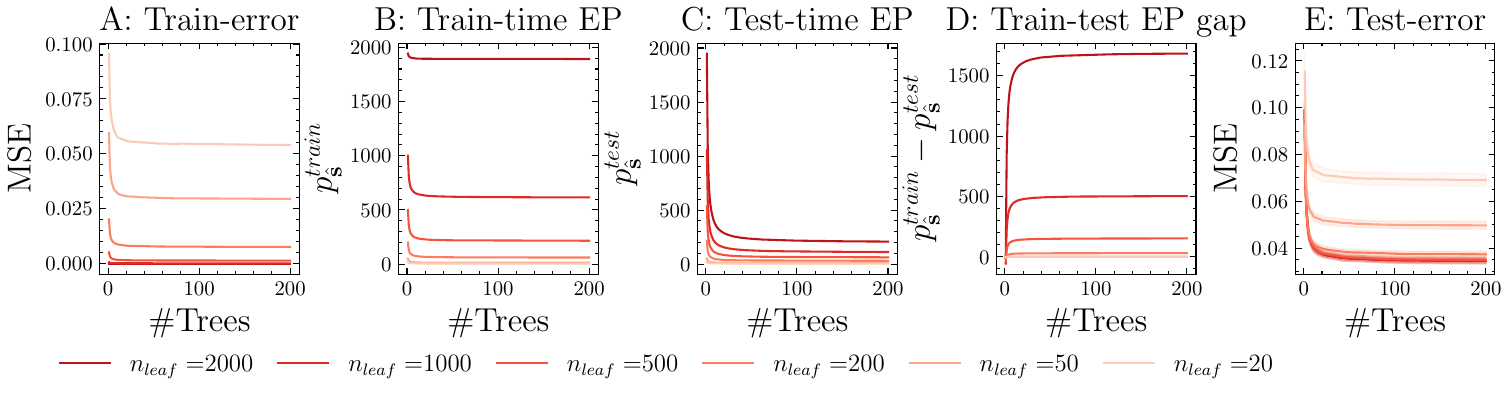}\vspace{-.3cm}
    \caption{\textbf{The smoothing effect of ensembling for trees of different depth on the MiamiHousing2016 dataset (Regression, $d=13$)} \small for forests of trees of different depths trained without bootstrapping and with $m=\frac{1}{3}$.}
    \label{fig:train-test-gap44147}
\end{figure}

\begin{figure}[h]
    \centering
\includegraphics[width=.99\textwidth]{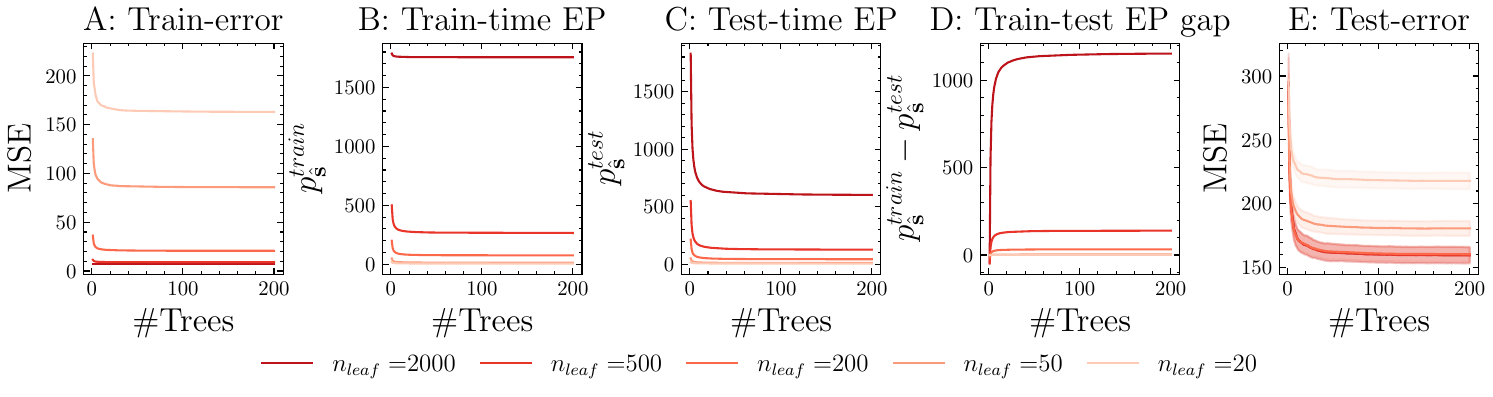}\vspace{-.3cm}
    \caption{\textbf{The smoothing effect of ensembling for trees of different depth on the Superconduct dataset (Regression, $d=79$)} \small for forests of trees of different depths trained without bootstrapping and with $m=\frac{1}{3}$.}
    \label{fig:train-test-gap44148}
\end{figure}

Next, we replicate the experiments from \cref{sec:spike-smooth} considering forests of trees of different depth as in \cref{fig:train-test-gap}. In \cref{fig:train-test-gap44147} and \cref{fig:train-test-gap44148}, we observe identical trends as in the main text for both datasets: when trees are not grown to full depth, ensembling also has a train-time smoothing effect.  Further, spiked-smooth behavior is not unique to interpolating models, but is much more pronounced when models are heavily overfitted. 

\begin{figure}
    \centering
\includegraphics[width=.99\textwidth]{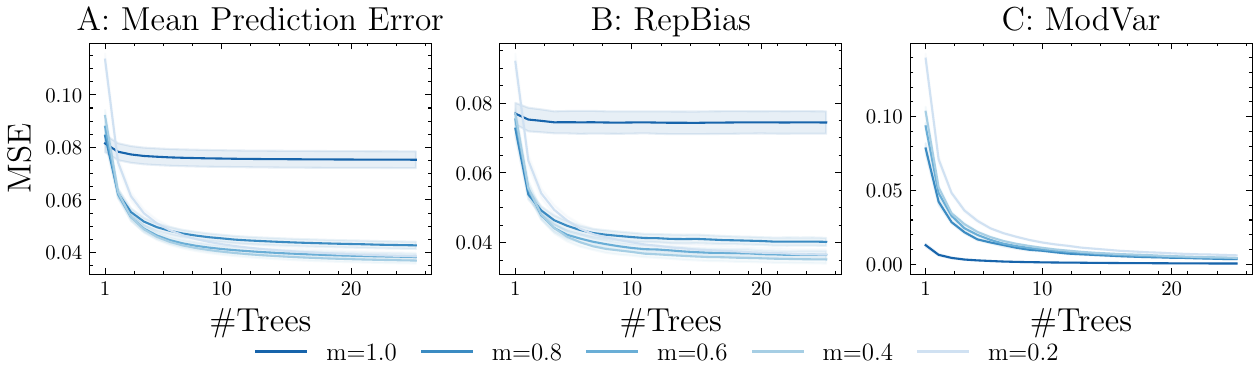}\vspace{-.3cm}
    \caption{\textbf{The effect of ensemble size on mean prediction error (A), representation bias (RepBias, B) and model variability (ModVar, C) on the MiamiHousing2016 dataset (Regression, $d=13$)} \small for interpolating trees. Evaluated by randomly re-initializing the same model 50 times and picking $\hat{\mu}^*_Z(\cdot)$ with oracle knowledge of the test outcomes. }
    \label{fig:error-decomp44147}
\end{figure}

\begin{figure}
    \centering
\includegraphics[width=.99\textwidth]{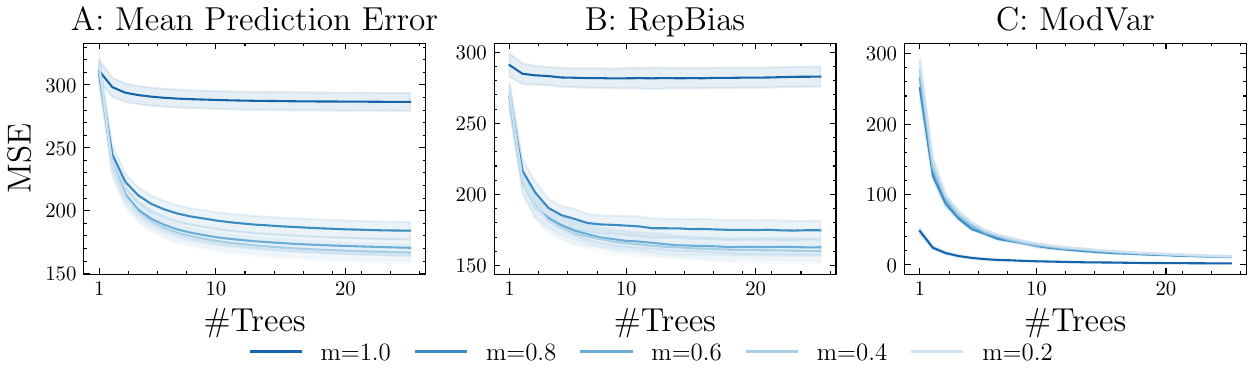}\vspace{-.3cm}
    \caption{\textbf{The effect of ensemble size on mean prediction error (A), representation bias (RepBias, B) and model variability (ModVar, C) on the Superconduct dataset (Regression, $d=79$)} \small for interpolating trees. Evaluated by randomly re-initializing the same model 50 times and picking $\hat{\mu}^*_Z(\cdot)$ with oracle knowledge of the test outcomes. }
    \label{fig:error-decomp44148}
\end{figure}

Finally, we replicate the experiment from \cref{sec:bias} and decompose observed error into RepBias and ModelVar as in \cref{fig:error-decomp}. In \cref{fig:error-decomp44147} and \cref{fig:error-decomp44148}, we indeed observe the same trends as in the main text: inducing randomness through $m<1$ appears to improve performance of an ensemble of trees relative to a single tree with $m=1$ mainly due to differences in what can be \textit{represented} in the best case by each model. 

\subsubsection{Classification datasets}
Next, we use two real binary classification datasets to replicate some of the analyses from the main text for classification outcomes. As a dataset with relatively few features, we use the California housing dataset (\url{https://www.openml.org/d/45028}, $n_{total}=20634$), in which the task is to predict whether house values in a district lie above the median from 8 features capturing district characteristics. As a second dataset, we use Bioresponse (\url{https://www.openml.org/d/45019}, $n_{total}=3434$), \cite{grinsztajn2022tree}'s classification dataset with most features, in which the task is to predict the response of molecules from 419 features capturing their chemical properties. As discussed in \cref{sec:treesassmoother}, we use trees that issue predictions by averaging (not voting) to stay consistent with the smoother formulation, which is also in line with recent work suggesting the use of the squared loss in classification \citep{hui2020evaluation, muthukumar2021classification}.

\begin{figure}
    \centering
\includegraphics[width=.99\textwidth]{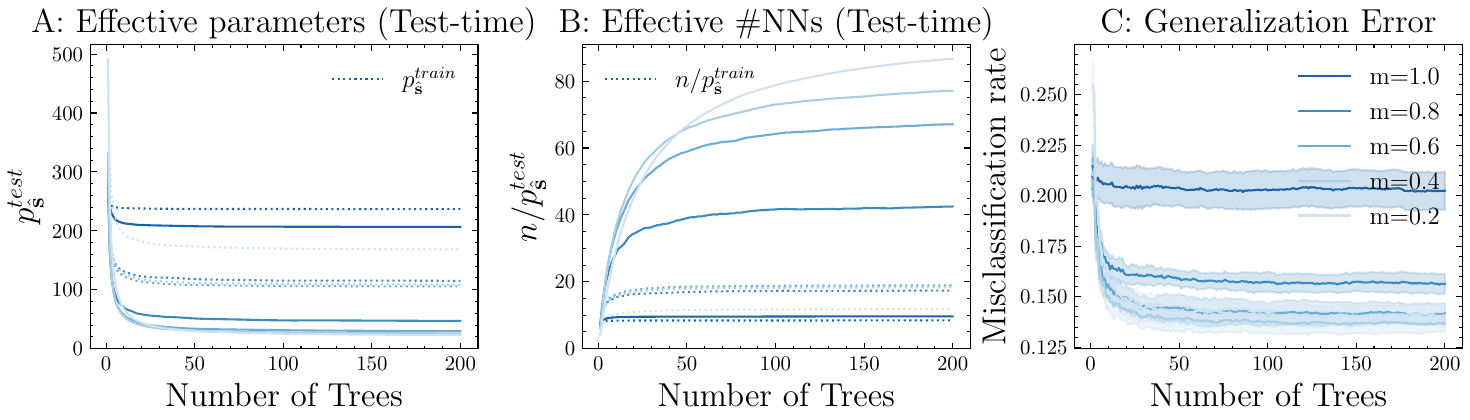}\vspace{-.3cm}
\caption{\textbf{Interpolating ensembles on the California housing dataset (Binary classification, $d=8$).} }
    \label{fig:wyner-45028}
\end{figure}

\begin{figure}
    \centering
    \includegraphics[width=.99\textwidth]{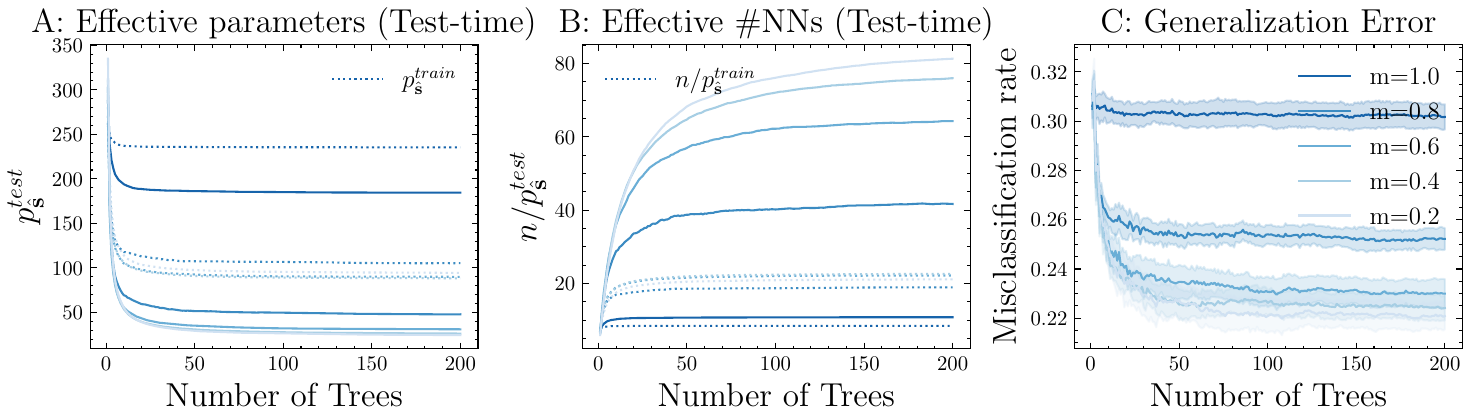}\vspace{-.3cm}
    \caption{\textbf{Interpolating ensembles on the Bioresponse dataset (Binary classification, $d=419$).} }
    \label{fig:wyner-45019}
\end{figure}

In \cref{fig:wyner-45028} and \cref{fig:wyner-45019}, we once more begin by replicating the experiment from \cref{sec:wynerexp}. Again, we find that ensembling of interpolating trees has the expected smoothing effect also on the test examples in this real data, and this indeed improves their generalization error as measured by the misclassification rate. Like for the real-world regression experiments, we find no real harmful effect of setting $m$ very small. The only notable difference in results appears to lie in (i) train-time effective parameters $\textstyle p^{train}_{\hat{\mathbf{s}}}$ now also lying far below $n_{train}=2000$ and (ii) the effective parameter numbers on the test examples also being much lower than before. Both are due to the fact that, unlike in regression, classification outcomes are \textit{not unique}, thus leaves will often be pure much before $n_{train}$ leaves are reached and thus tree construction will stop before full-depth is reached. That is, interpolating classification trees will often need to be much less deep than interpolating regression trees.

\begin{figure}[!t]
    \centering
\includegraphics[width=.99\textwidth]{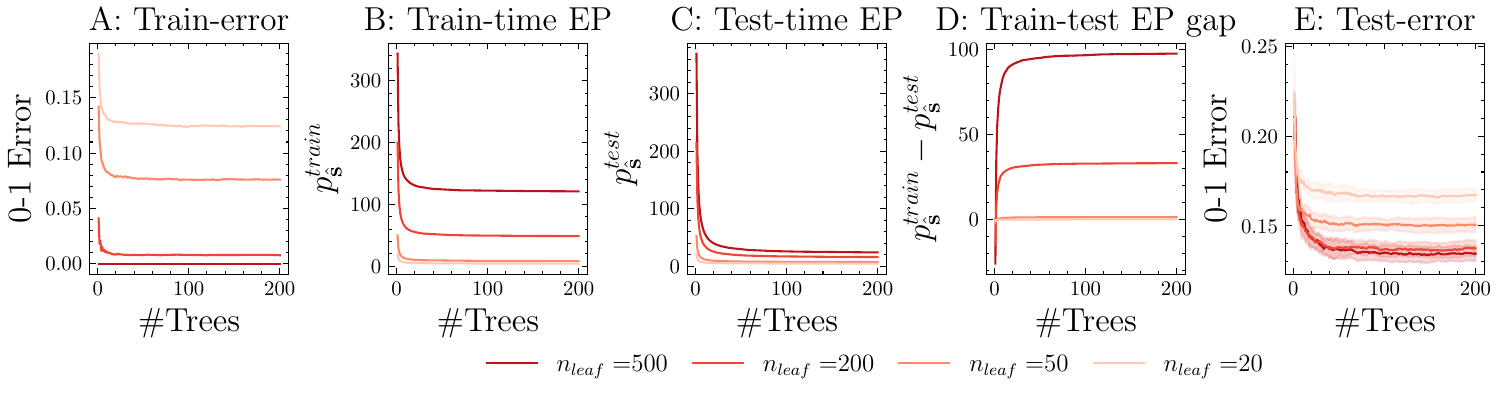}\vspace{-.3cm}
    \caption{\textbf{The smoothing effect of ensembling for trees of different depth on the California housing dataset (Binary classification, $d=8$)} \small for forests of trees of different depths trained without bootstrapping and with $m=\frac{1}{3}$.}
    \label{fig:train-test-gap45028}
\end{figure}

\begin{figure}[!t]
    \centering
\includegraphics[width=.99\textwidth]{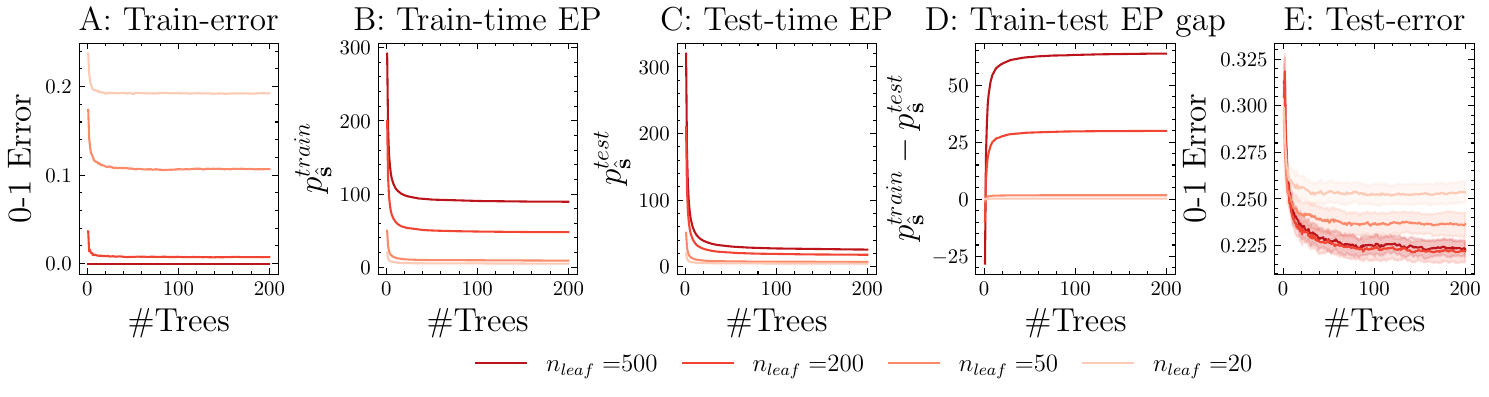}\vspace{-.3cm}
    \caption{\textbf{The smoothing effect of ensembling for trees of different depth on the Bioresponse dataset (Binary classification, $d=419$)} \small for forests of trees of different depths trained without bootstrapping and with $m=\frac{1}{3}$.}
    \label{fig:train-test-gap45019}
\end{figure}

This is precisely what we observe when replicating the experiments from  \cref{sec:spike-smooth} considering forests of trees of different depth as in \cref{fig:train-test-gap}. In \cref{fig:train-test-gap45028} and \cref{fig:train-test-gap45019}, we observe that the training data is already interpolated when using only 500 leaves. The results remain consistent with what was observed in the regression experiments.

\begin{figure}[!b]
    \centering
\includegraphics[width=.99\textwidth]{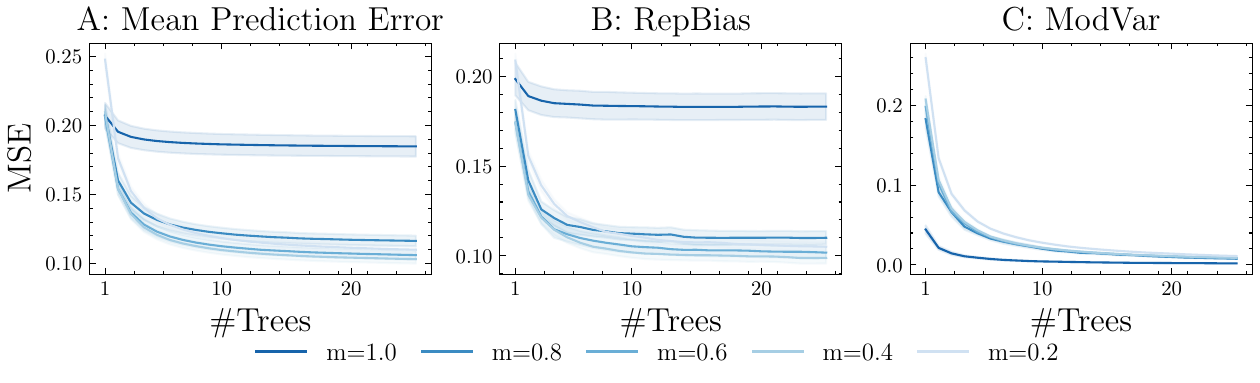}\vspace{-.3cm}
    \caption{\textbf{The effect of ensemble size on mean prediction error (A), representation bias (RepBias, B) and model variability (ModVar, C) on the California housing dataset (Binary classification, $d=8$)} \small for interpolating trees. Evaluated by randomly re-initializing the same model 50 times and picking $\hat{\mu}^*_Z(\cdot)$ with oracle knowledge of the test outcomes. }
    \label{fig:error-decomp45028}
\end{figure}

\newpage
Finally, we replicate the results from \cref{sec:bias}. As we defined RepBias and ModVar with respect to the squared loss, we will report squared test loss here. In \cref{fig:error-decomp45028} and \cref{fig:error-decomp45028} we find that our findings from the main text also replicate on these two classification datasets.

\begin{figure}
    \centering
\includegraphics[width=.99\textwidth]{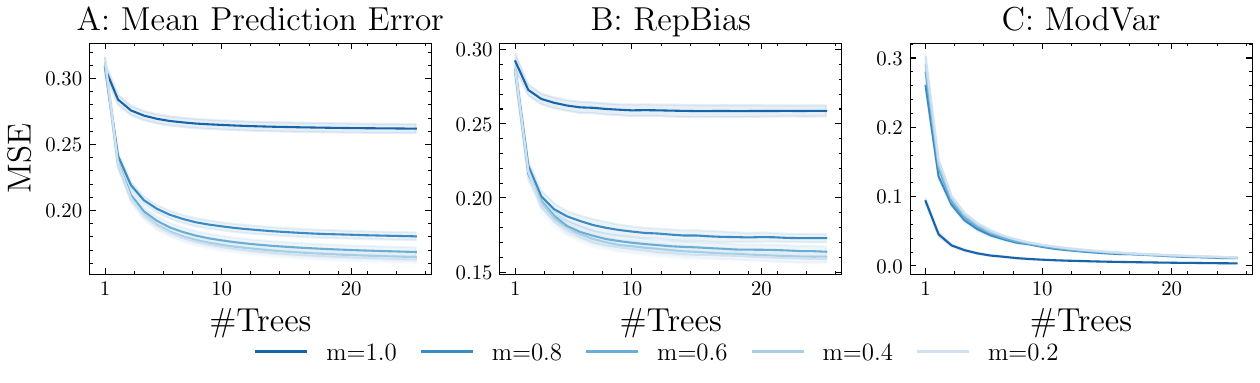}\vspace{-.3cm}
    \caption{\textbf{The effect of ensemble size on mean prediction error (A), representation bias (RepBias, B) and model variability (ModVar, C) on the Bioresponse dataset (Binary classification, $d=419$)} \small for interpolating trees. Evaluated by randomly re-initializing the same model 50 times and picking $\hat{\mu}^*_Z(\cdot)$ with oracle knowledge of the test outcomes. }
    \label{fig:error-decomp45019}
\end{figure}

\subsection{Additional experiments on simulated data}
\begin{figure}[!b]
    \centering
\includegraphics[width=.99\textwidth]{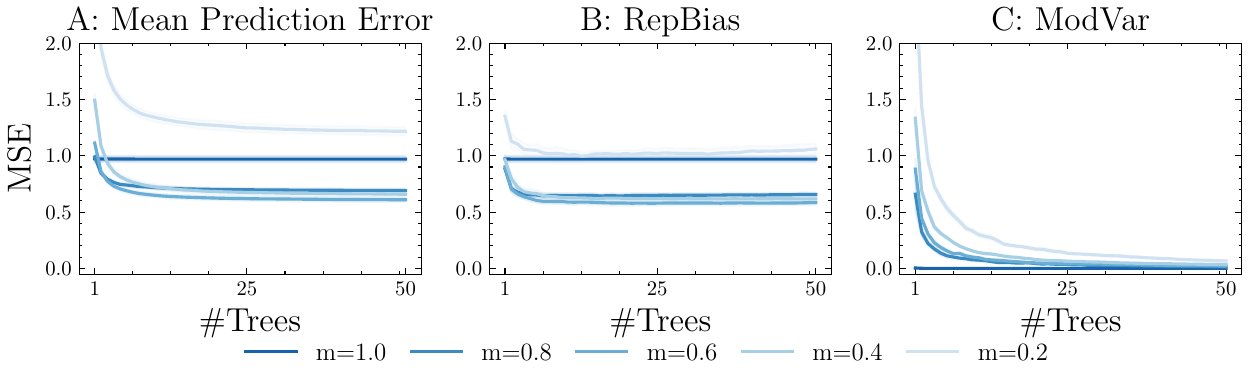}\vspace{-.3cm}
    \caption{\textbf{The effect of ensemble size on mean prediction error (A), representation bias (RepBias, B) and model variability (ModVar, C)} on the MARSAdd simulation\small for  trees with 20 leaves. Evaluated in a setting without label noise ($\sigma=0$) and with intermediate train-test dissimilarity ($\delta=0.1$) by randomly re-initializing the same model 50 times and picking $\hat{\mu}^*_Z(\cdot)$ with oracle knowledge of the test outcomes.}
    \label{fig:error-decomp-small}
\end{figure}

Here, we demonstrate that our findings regarding the effect of ensembling on RepBias and ModVar presented in \cref{fig:error-decomp} replicate when using non-interpolating trees. Instead of using full-depth (interpolating) trees with $500=n_{train}$ leaves as in the main text, we replicate the findings using much shallower trees of 20 leaves in \cref{fig:error-decomp-small} -- and observe qualitatively identical trends.

\subsection{Implementation details}
We use regression trees and random forests as implemented in \texttt{RandomForestRegressor} in scikit-learn \citep{scikit-learn} for all experiments, which in turn implement an optimized version of the CART algorithm \citep{breiman1984classification}. For the boosting experiments,  we rely on the scikit-learn implementation \texttt{GradientBoostingRegressor}.

\end{document}